\documentclass[a4paper,draftcls,12pt,onecolumn,journal]{IEEEtran}
\ifCLASSINFOpdf
\else
\fi
\usepackage[cmex10]{amsmath}
\usepackage{amsthm}
\usepackage{multirow,tabularx}
\usepackage{setspace,lscape,rotating} 
\usepackage[CaptionAfterwards]{fltpage}

\theoremstyle{remark}

\usepackage{algorithm,latexsym,mathrsfs,graphicx}
\usepackage{algpseudocode}

\usepackage{ccaption}

\usepackage[normalem]{ulem}
\usepackage[font=normalsize,labelfont=sf,textfont=sf]{caption}
\DeclareCaptionLabelFormat{andtable}{#1~#2  \&  \tablename~\thetable}
\AtBeginCaption{\doublespacing}

\usepackage{array}

\ifCLASSOPTIONcompsoc
  \usepackage[caption=false,font=normalsize,labelfont=sf,textfont=sf]{subfig}
\else
  \usepackage[caption=false,font=normalsize,labelfont=sf,textfont=sf]{subfig}
\fi

\begin{document}

\title{On the convergence of the sparse possibilistic c-means algorithm} 
%
%
%
%

\DeclareRobustCommand*{\IEEEauthorrefmark}[1]{%
  \raisebox{0pt}[0pt][0pt]{\textsuperscript{\footnotesize\ensuremath{#1}}}}

\author{\IEEEauthorblockN{Konstantinos~D.~Koutroumbas\IEEEauthorrefmark{1}, Spyridoula~D.~Xenaki\IEEEauthorrefmark{1,}\IEEEauthorrefmark{2}, and
Athanasios~A.~Rontogiannis\IEEEauthorrefmark{1}}

\IEEEauthorblockA{\IEEEauthorrefmark{1}{\normalsize Institute for Astronomy, Astrophysics, Space Applications and Remote Sensing (IAASARS), National Observatory of Athens, Penteli, GR-15236 Greece}}\\
\IEEEauthorblockA{\IEEEauthorrefmark{2}{\normalsize Department of Informatics and Telecommunications, National \& Kapodistrian University of Athens, GR-157 84, Ilissia, Greece}}}
\IEEEtitleabstractindextext{%
\begin{abstract}
In this paper, a convergence proof for the recently proposed cost function optimization sparse possibilistic c-means (SPCM) algorithm is provided. Specifically, it is shown that the algorithm will converge to one of the local minima of its associated cost function. It is also shown that similar convergence results can be derived for the well-known possibilistic c-means (PCM) algorithm proposed in \cite{Kris96}, if we view it as a special case of SPCM. Note that the convergence results for PCM are stronger than those established in previous works.

\end{abstract}

}

\maketitle

\IEEEdisplaynontitleabstractindextext

%
\IEEEpeerreviewmaketitle

\section{Introduction}

\IEEEPARstart{I}{n} most of the well-known clustering algorithms that deal with the identification of compact and hyperellipsoidally shaped clusters, each cluster is represented by a vector called {\it cluster representative} that lie in the same feature space with the data vectors. In order to identify the underlying clustering structure, such algorithms gradually move the representatives from their initial (usually randomly selected) locations towards the ``center" of each cluster. Apart from hard clustering philosophy, where each data vector belongs exclusively to a single cluster (e.g. {\it k-means} \cite{Hart79}) and fuzzy clustering philosophy, where each data vector is shared among the clusters (e.g. {\it fuzzy c-means} (FCM) \cite{Bezd80}, \cite{Bezd81}), an alternative well-known clustering philosophy that has been developed, in order to deal with this case, is the possibilistic clustering one, where the {\it degree of compatibility} of a data vector with a given cluster is independent of its degrees of compatibility with any other cluster. Algorithms of this kind, known as possibilistic c-means algorithms (PCMs), iteratively optimize suitably defined cost functions (e.g. \cite{Kris93}, \cite{Kris96}, \cite{Pal05}, \cite{Yang06}, \cite{Tree13}, \cite{Theo09}), aiming at moving the cluster representatives to regions that are dense in data points. A very well-known PCM algorithm, introduced in \cite{Kris93} and noted as PCM$_1$, is derived from the minimization of the cost function 
\begin{equation}
J_{PCM_1}(U,\Theta)=\sum\limits_{i=1}^N \sum\limits_{j=1}^m u_{ij}^q\|\mathbf{x}_i-\boldsymbol{\theta}_j\|^2 + \sum\limits_{j=1}^m \gamma_j \sum\limits_{i=1}^N (1-u_{ij})^q,
\label{Jpcm1}
\end{equation}
while an alternative PCM algorithm, presented in \cite{Kris96} and noted as PCM$_2$, is derived from the minimization of the cost function
\begin{equation}
J_{PCM_2}(U,\Theta)=\sum\limits_{i=1}^N \sum\limits_{j=1}^m u_{ij}\|\mathbf{x}_i-\boldsymbol{\theta}_j\|^2 + \sum\limits_{j=1}^m \gamma_j \sum\limits_{i=1}^N (u_{ij}\ln u_{ij}-u_{ij})
\label{Jpcm2}
\end{equation}
where $\mathbf{x}_i$, $i=1,\ldots,N$ denotes the $i$th out of $N$ $l$-dimensional data points of the data set $X$ under study, $\boldsymbol{\theta}_j$'s, $j=1,\ldots,m$ denote the representatives of the $m$ clusters (each one denoted by $C_j$), which constitute the set $\Theta$. $U$ is the matrix, whose $(i, j)$ element $u_{ij}$ stands for the {\it degree of compatibility} of the $i$th data vector $\mathbf{x}_i$ with the $j$th representative $\boldsymbol{\theta}_j$. Finally, $\gamma_j$'s are positive parameters, each one associated with a cluster $C_j$ \footnote{Note that, in contrast to $J_{PCM_2}$,  $J_{PCM_1}$ involves an additional parameter $q$, which takes values around $2$.}.

Convergence results of these algorithms have been presented, utilizing the Zangwill convergence theorem \cite{Zang69}. It is shown that the iterative sequence generated by a PCM converges to either (a) a local minimizer or a saddle point of the cost function associated with the algorithm or (b) any of its convergent subsequences converges to either a local minimizer or a saddle point of the cost function \cite{Zhou13}. It is noteworthy that Zangwill's theorem \cite{Zang69} has been used to establish convergence properties for the FCM algorithm as well (e.g. \cite{Bezd80}, \cite{Bezd87}, \cite{Hath87})\footnote{A different approach for proving the convergence of the FCM to a stationary point of the corresponding cost function is given in \cite{Grol05}. A relative work is also provided in \cite{Hopp03}.}.

Recently, a novel possibilistic clustering algorithm, called Sparse Possibilistic C-Means (SPCM) \cite{Xena16}, has been proposed, which extends PCM$_2$ by introducing sparsity. More specifically, a suitable sparsity constraint is imposed on the vectors containing the degrees of compatibility of the data points with the clusters (one vector per point\footnote{Clearly, these vectors are the rows of the matrix $U$.}), such that each data vector is compatible with only a few or even {\it none} clusters. In the present work, an analysis of the convergence properties of SPCM algorithm is conducted and it is shown that the iterative sequence generated by SPCM converges to a local minimum of its associated cost function $J_{SPCM}$, which is defined explicitly in the next section. A significant source of difficulties in the convergence analysis of SPCM is the addition of an extra term in the cost function $J_{{PCM}_2}$, as explained in the next section, that is responsible for sparsity imposition, which gives the main novelty of SPCM. This affects the updating of the degrees of compatibility, which now are not given in closed form and they are computed via a two-branch expression. 

Moreover, it is shown that the above convergence analysis for SPCM is directly applicable to the PCM$_2$ algorithm (\cite{Kris96}) and the obtained convergence results are much stronger than those provided in \cite{Zhou13}.

The rest of the paper is organized as follows. In Section II, a brief description of the SPCM algorithm is given for reasons of thoroughness and in Section III its convergence proof is analyzed. In Section IV the convergence results from the previous section are applied for the case of PCM$_2$. Finally, Section V concludes the paper.

\section{The Sparse PCM (SPCM) algorithm}

Let $X=\{\mathbf{x}_i \in {\cal R}^l, i=1,...,N\}$ be the data set under study, $\Theta=\{\boldsymbol{\theta}_j \in {\cal R}^l, j=1,...,m\}$ be a set of $m$ vectors that will be used for the representation of the clusters formed in $X$ ({\it cluster representatives}) and $U=[u_{ij}], i=1,...,N, j=1,...,m$ be an {$N\times m$} matrix whose $(i,j)$ element stands for the {\it degree of compatibility} of $\mathbf{x}_i$ with the $j$th cluster. Let also ${\mathbf{u}_i}^T=[u_{i1},...,u_{im}]$ be the (row) vector containing the elements of the $i$th row of $U$. In what follows we consider only Euclidean norms, denoted by $\|\cdot\|$.

As it has been stated earlier, the strategy of a possibilistic algorithm is to move the vectors $\boldsymbol{\theta}_j$'s towards regions that are dense in data points of $X$ (clusters). The aim of SPCM is two-fold: (a) to retain the sparser clusters, provided of course that at least one representative has been initially placed in each one of them and (b) to prevent noisy points from contributing to the computation of any of the $\boldsymbol{\theta}_j$'s. This is achieved by suppressing the contribution of data points that are distant from a representative $\boldsymbol{\theta}_j$ in its updating. More specifically, focusing on a specific representative $\boldsymbol{\theta}_j$, this can be achieved by setting $u_{ij}=0$ for data points $\mathbf{x}_i$ that are distant from it. This is tantamount to imposing sparsity on $\mathbf{u}_i$, i.e., forcing the corresponding data point $\mathbf{x}_i$ to contribute only to its (currently) closest representatives. To this end, the cost function $J_{PCM_2}$ of eq.~\eqref{Jpcm2} is augmented as follows,
\begin{equation}
J_{SPCM}(U,\Theta)=\sum_{j=1}^m \left[ \sum_{i=1}^N u_{ij} \|\mathbf{x}_i-\boldsymbol{\theta}_j\|^2 + \gamma_j \sum_{i=1}^N (u_{ij} \ln u_{ij} - u_{ij})\right] + \lambda \sum_{i=1}^N \|\mathbf{u}_i\|_p^p, \ u_{ij}> 0 \ \footnote{This is a prerequisite in order for the $\ln u_{ij}$ to be well-defined. However, in the sequel, when refering to $\ln u_{ij}$ for $u_{ij}=0$, we mean $\lim\limits_{u_{ij}\rightarrow 0^+}u_{ij}$. Also, we use the fact that $\lim\limits_{u_{ij}\rightarrow 0^+}u_{ij}\ln u_{ij}=0$.},\label{Jspcm}
\end{equation}
where $\|\mathbf{u}_i\|_p$ is the $\ell_p$-norm of vector $\mathbf{u}_i$ ($p \in (0, 1)$); thus, $\|\mathbf{u}_i\|_p^p=\sum_{j=1}^m u_{ij}^p$. Each $\gamma_j$ indicates the degree of ``influence" of $C_j$ around its representative $\boldsymbol{\theta}_j$; the smaller (greater) the value of $\gamma_j$, the smaller (greater) the influence of cluster $C_j$ around $\boldsymbol{\theta}_j$. The last term in eq.~(\ref{Jspcm}) is expected to induce sparsity on each one of the vectors $\mathbf{u}_i$ and $\lambda$ ($\geq 0$) is a regularization parameter that controls the degree of the imposed sparsity. The algorithm resulting by the minimization of $J_{SPCM}(U,\Theta)$ is called {\it sparse possibilistic c-means} (SPCM) clustering algorithm and it is briefly discussed below (its detailed presentation is given in \cite{Xena16}).

\subsection{Initialization in SPCM}
First, the initialization of $\boldsymbol{\theta}_j$'s is carried out using the final cluster representatives obtained from the FCM algorithm, when the latter is executed with $m$ clusters on $X$.

After the initialization of $\boldsymbol{\theta}_j$'s, we initialize $\gamma_j$'s as follows:
\begin{equation}
\label{initgamma}
\gamma_j=\frac{ \sum_{i=1}^N u^{FCM}_{ij} \|\mathbf{x}_i-\boldsymbol{\theta}_j\|^2}{ \sum_{i=1}^N u^{FCM}_{ij} }, \ \ \ j=1,\ldots,m
\end{equation}
where $\boldsymbol{\theta}_j$'s and $u^{FCM}_{ij}$'s in eq.~(\ref{initgamma}) are the final parameter estimates obtained by FCM. 

Finally, we select the parameter $\lambda$ as follows:
\begin{equation}
\lambda=K\frac{\bar{\gamma}}{p(1-p)\mathrm{e}^{2-p}},
\label{lambda}
\end{equation}
where $\bar{\gamma}=\min\limits_{j=1,\ldots,m}\gamma_j$ and $K$ is a user-defined constant, which is set equal to $K=0.9$ for $p=0.5$ (see also \cite{Xena16}). The rationale behind this choice is further enlightened in subsection \ref{subsecKpara}, where, in addition, appropriate bounds on the values of $K$ are given in terms of $p$.


\subsection{Updating of $\boldsymbol{\theta}_j$'s and $u_{ij}$'s in SPCM}
Minimizing $J_{SPCM}(U,\Theta)$ with respect to $\boldsymbol{\theta}_j$ leads to the following equation,
\begin{equation}
	\boldsymbol{\theta}_j=\frac{\sum_{i=1}^N u_{ij} \mathbf{x}_i}{\sum_{i=1}^N u_{ij}}
\label{theta}
\end{equation}
%
%
%
%
%
%
%

\begin{figure}[]
\centering
\subfloat[$f(u_{ij})$]{\includegraphics[width=0.4\textwidth]{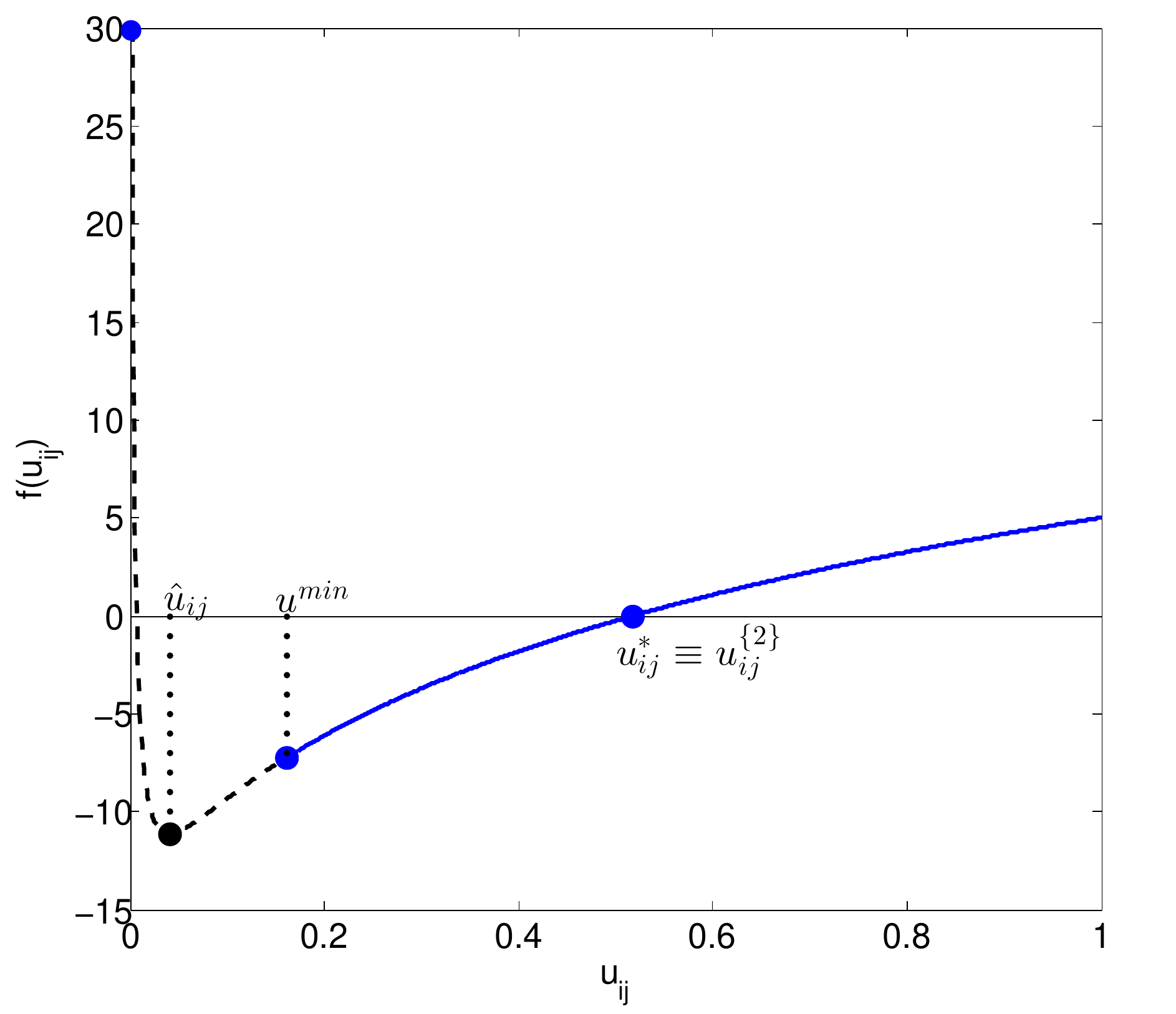}\label{f1}}
\hfil
\centering
\subfloat[$J(u_{ij})$]{\includegraphics[width=0.4\textwidth]{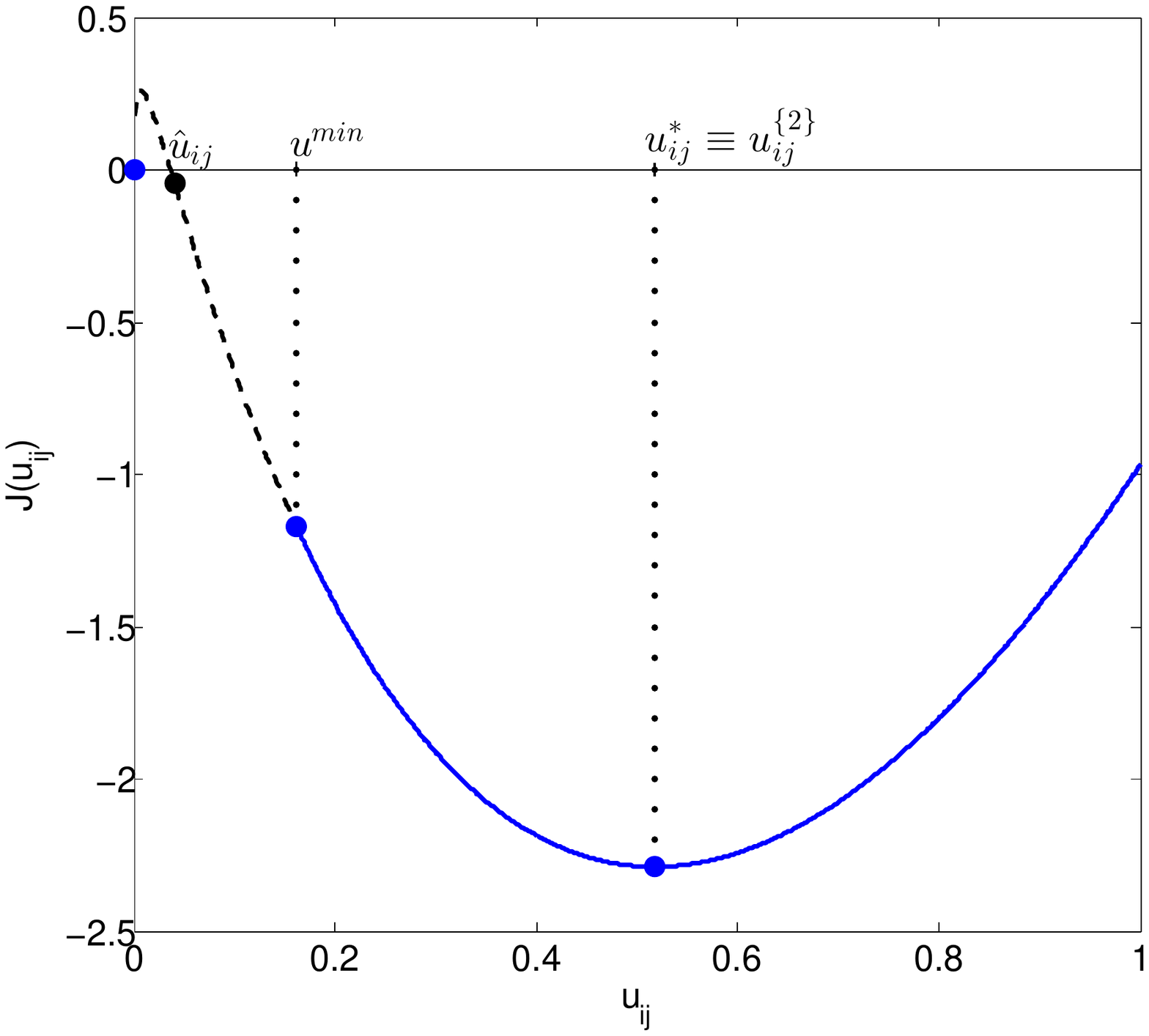}\label{J1}}
\hfil
\centering
\subfloat[$f(u_{ij})$]{\includegraphics[width=0.4\textwidth]{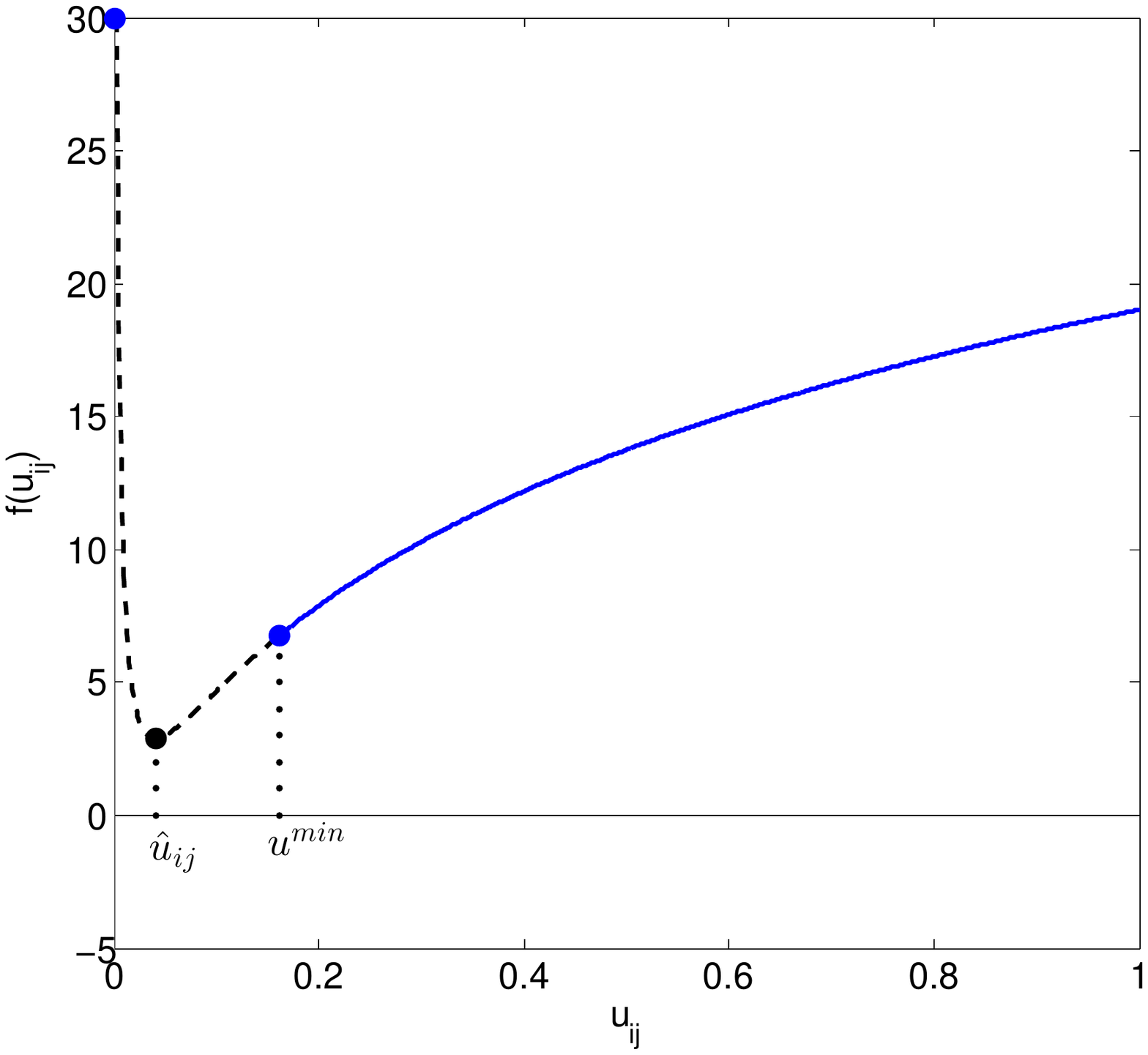}\label{f2}}
\hfil
\centering
\subfloat[$J(u_{ij})$]{\includegraphics[width=0.4\textwidth]{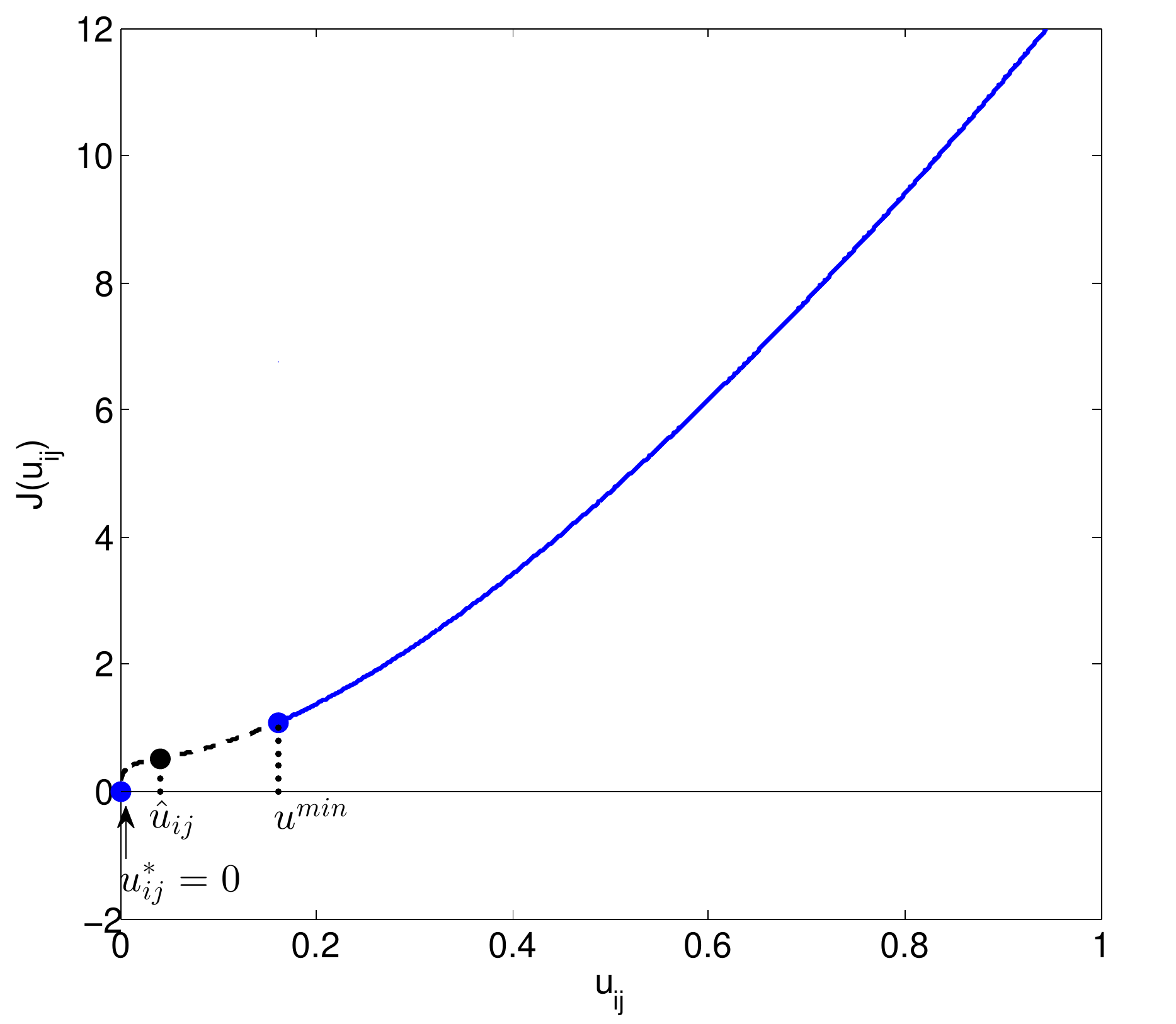}\label{J2}}
\hfil
\centering
\subfloat[$f(u_{ij})$]{\includegraphics[width=0.4\textwidth]{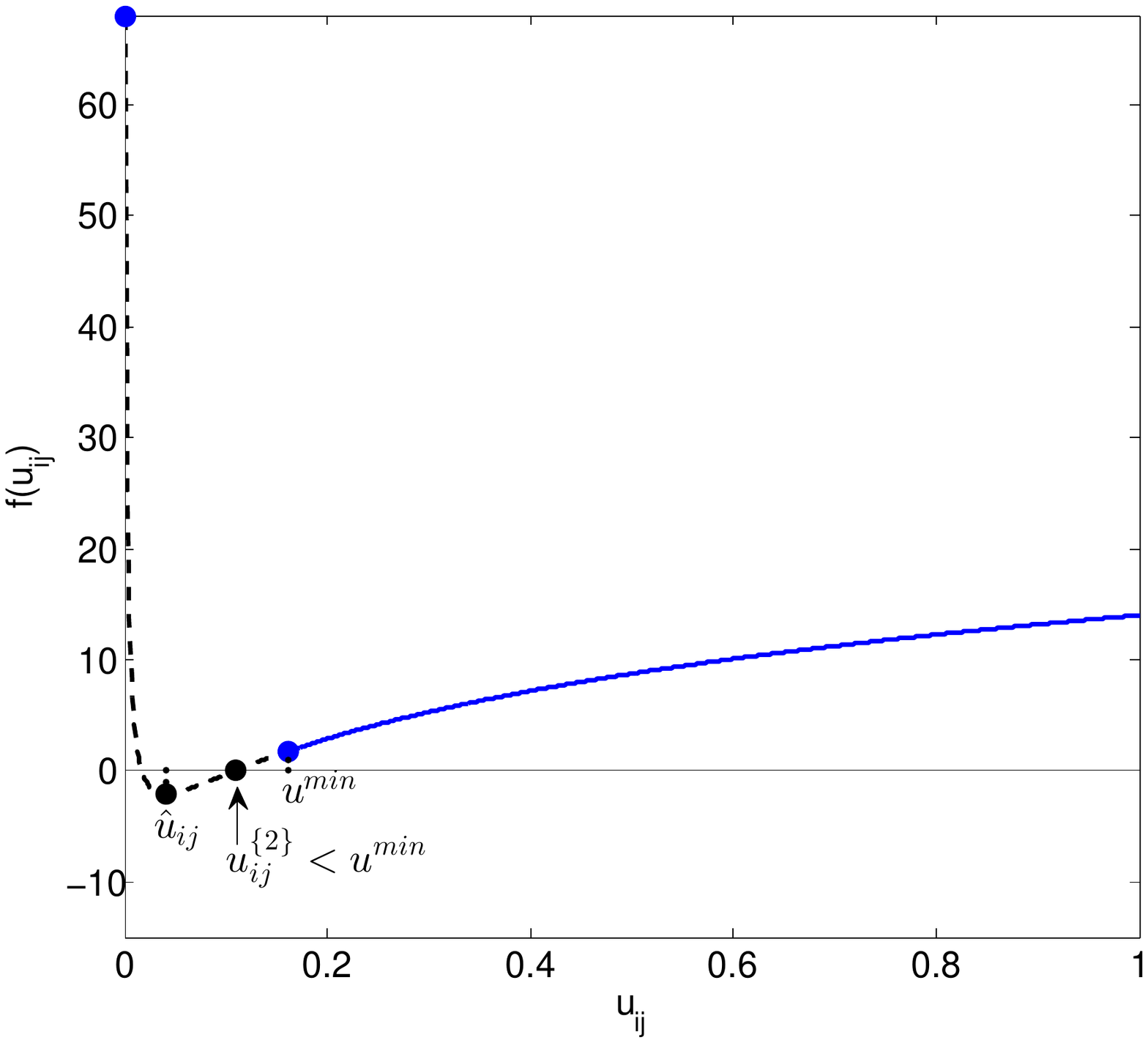}\label{f3}}
\hfil
\centering
\subfloat[$J(u_{ij})$]{\includegraphics[width=0.4\textwidth]{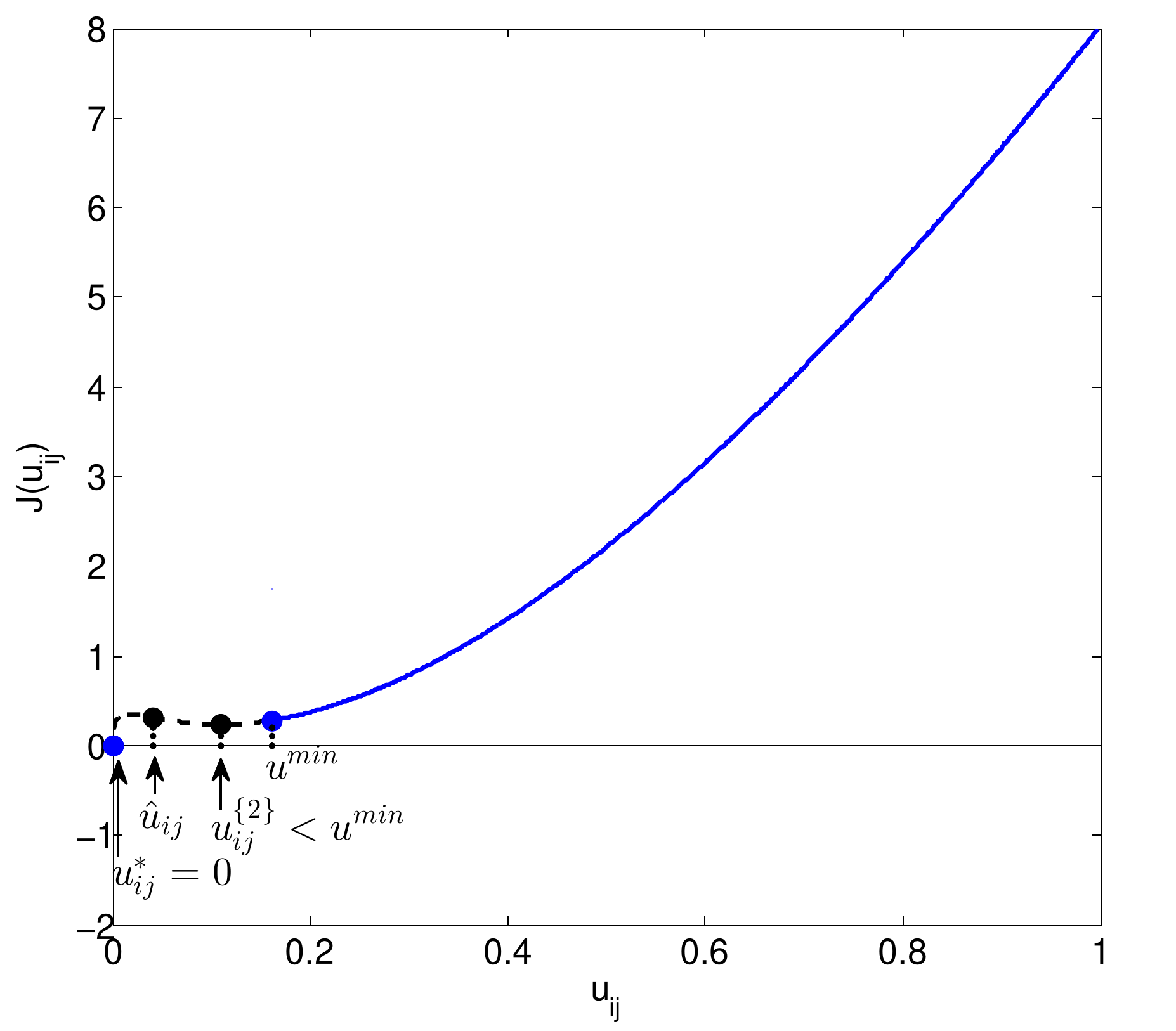}\label{J3}}
\hfil
\centering{\caption{\small\onehalfspacing In all plots the dashed parts of the graphs correspond to the interval $(0,u_{min})$, which is not accessible by the algorithm (see eq.~\eqref{globJ}). (a) The shape of function $f(u_{ij})$, when $f(\hat{u}_{ij})<0$ and the right-most condition of eq.~\eqref{globJ} is satisfied and (b) the corresponding shape of the cost function $J(u_{ij})$. (c) The shape of function $f(u_{ij})$, when $f(\hat{u}_{ij})>0$ and (d) the corresponding shape of $J(u_{ij})$. (e) The shape of function $f(u_{ij})$, when $f(\hat{u}_{ij})<0$ and the right-most condition of eq.~\eqref{globJ} is not satisfied and (f) the corresponding shape of $J(u_{ij})$.}\label{fJplots}}
\end{figure}

The derivative of $J_{SPCM}$ with respect to $u_{ij}$ is $f(u_{ij})=d_{ij}+\gamma_j \ln u_{ij} + \lambda p u_{ij}^{p-1}$, where $d_{ij}=\|\mathbf{x}_i-\boldsymbol{\theta}_j\|^2$. In \cite{Xena16} it is proved that (a) $f(u_{ij})$ is strictly positive outside $[0,\ 1]$, (b) $f(u_{ij})$ has a unique minimum at $\hat{u}_{ij}=[ \frac{\lambda}{\gamma_j} p(1-p)]^\frac{1}{1-p}$ and (c) $f(u_{ij})=0$ has at most two solutions. More specifically, if $f(\hat{u}_{ij})<0$ , then $f(u_{ij})=0$ has exactly two solutions $u_{ij}^{ \{1\} }, u_{ij}^{ \{2\} } \in (0,\ 1)$, with $u_{ij}^{ \{1\} } < u_{ij}^{ \{2\} }$, the largest of which corresponds to a local minimum of $J_{SPCM}$ with respect to $u_{ij}$. In \cite{Xena16} it is shown that $J_{SPCM}(U,\Theta)$ exhibits its global minimum at $u^*_{ij}$, where: 
\begin{equation}
\label{globJ}
u^*_{ij}=\left\{\begin{matrix}
u^{\{2\}}_{ij}, & \text{if } f(\hat{u}_{ij})<0 \text{ and } u^{\{2\}}_{ij} \geq \left(\frac{\lambda(1-p)}{\gamma_j}\right)^{1/(1-p)} (\equiv u^{min})\\ 
0, & \text{otherwise} \ \ 
\end{matrix}\right. \ \ \ \footnote{In its original version, the second inequality in the first branch was strict. Here, we change it to ``less than or equal to''. Although this slight modification has no implications to the behavior of the algorithm in practice, it turns out to be important for the establishment of the theoretical results given below.}
\end{equation} 
%
Clearly, if $f(u_{ij})=0$ has no solutions, then $f(u_{ij})$ will be positive for all valid values of $u_{ij}$ (see Fig.~\ref{f2}). Thus $J_{SPCM}$ will be strictly increasing and it will be minimized at 0. Thus, we set $u^*_{ij}=0$. Note that the right-most inequality in the first branch of eq.~\eqref{globJ} turns out to be equivalent to $J_{SPCM}(\boldsymbol{\theta}_j,u_{ij}^{\{2\}}) \leq J_{SPCM}(\boldsymbol{\theta}_j,0)=0$, where $J_{SPCM}(\boldsymbol{\theta}_j,u_{ij})$ contains the terms of $J_{SPCM}(U,\Theta)$ that involve only $\boldsymbol{\theta}_j$ and $u_{ij}$ (\cite{Xena16}). All the above possible cases are depicted in Fig.~\ref{fJplots}.

%

To determine $u_{ij}^{*}$, we solve $f(u_{ij})=0$ as follows. First, we determine $\hat{u}_{ij}$ and check whether $f(\hat{u}_{ij})>0$. If this is the case, then $f(u_{ij})$ has no roots in $[0,1]$. Note that, in this case, it is $f(u_{ij})>0$ for all $u_{ij}\in (0, 1]$, since $f(\hat{u}_{ij})>0$ (see Fig.~\ref{f2}). Thus, $J_{SPCM}$ is increasing with respect to $u_{ij}$ in $(0, 1]$ (see Fig.~\ref{J2}). Consequently, in this case we set $u_{ij}^{*}=0$, {\it imposing sparsity}. In the rare case, where $f(\hat{u}_{ij})=0$, we set $u_{ij}^{*}=0$, as $\hat{u}_{ij}$ is the unique root of $f(u_{ij})=0$ and $f(u_{ij})>0$ for $u_{ij}\in (0,\hat{u}_{ij})\cup(\hat{u}_{ij},1]$. If $f(\hat{u}_{ij})<0$, then $f(u_{ij})=0$ has exactly two solutions that both lie in $[0,1]$ (see Figs.~\ref{f1},~\ref{f3}). In order to determine the largest of the solutions ($u^{\{2\}}_{ij}$), we apply the bisection method (see e.g. \cite{Corl77}) in the range $(\hat{u}_{ij},1]$, as $u^{\{2\}}_{ij}$ is greater than $\hat{u}_{ij}$. The bisection method is known to converge very rapidly to the optimum $u_{ij}$, that is, in our case, to the largest of the two solutions of $f(u_{ij})=0$. If the obtained solution $u_{ij}^{\{2\}}$ satisfies the rightmost condition in the first branch of eq.~\eqref{globJ}, then we set $u_{ij}^{*}=u_{ij}^{\{2\}}$ (see Fig.~\ref{J1}), as is shown in \cite{Xena16}. Otherwise, $u_{ij}^{*}$ is set to 0 (see Fig.~\ref{J3}).

A vital observation is that, as long as $u_{ij}$ is given by the first branch of eq. (\ref{globJ}), its values are bounded as follows
\begin{equation}
\label{bound-u}
u^{min} \leq u_{ij} \leq u^{max}
\end{equation}
where $u^{max}$ is obtained by solving the equation $f(u_{ij})=0$, for $d_{ij}=0$; that is the equation $\gamma_j \ln u_{ij} + \lambda p u_{ij}^{p-1}=0$. Note that both $u^{min}$ and $u^{max}$ depend exclusively on $\lambda$, $\gamma_j$ and $p$.

Before we proceed, we will give an alternative expression for eq. (\ref{globJ}), which will be extensively exploited in the convergence proof below. More specifically, we will express the condition of the first branch of (\ref{globJ}) in terms of $\boldsymbol{\theta}_j$. To this end, we consider the case where $u_{ij}^{ \{ 2\} }=u^{min}$. This implies that $f(u_{ij}^{ \{ 2\} })=0$ or $f(u^{min})=0$. Substituting $u^{min}$ by its equal given in eq. (\ref{globJ}) and after some straightforward algebraic manipulations, it follows that $f(u_{ij}^{min})=0$ is equivalent to
\begin{equation}
\label{R2}
||\mathbf{x}_i - \boldsymbol{\theta}_j||^2 = \overbrace{ \frac{\gamma_j}{1-p} \left( -\ln \frac{\lambda(1-p)}{\gamma_j} -p \right)}^{R_j^2}
\end{equation}
The above is the equation of a hypersphere, denoted by ${\cal C}_{ij}$, centered at $\mathbf{x}_i$ and having radius $R_j$ (note that $R_j$ depends exclusively on the parameters $\gamma_j$, $p$, $\lambda$ and not on the data points $\mathbf{x}_i$ or on $\boldsymbol{\theta}_j$'s and $u_{ij}$'s). Clearly, its interior $int({\cal C}_{ij})$ (which in the subsequent analysis is assumed to contain ${\cal C}_{ij}$ itself) contains all the positions of $\boldsymbol{\theta}_j$ which give $u_{ij}>0$, while all the points in its exterior $ext({\cal C}_{ij})$ corresponds to positions of $\boldsymbol{\theta}_j$ that give $u_{ij}=0$. In order to ensure that ${\cal C}_{ij}$ is properly defined, we should ensure that $R_j$ is positive. This holds true if $K$ is chosen so that $K<p e^{2(1-p)}$ (see Proposition A1 in Appendix).
In the light of the above result, eq. (\ref{globJ}) can be rewritten as follows
\begin{equation}
\label{globJ1}
u^*_{ij}=\left\{\begin{matrix}
u^{\{2\}}_{ij}, & \text{if } ||\mathbf{x}_i - \boldsymbol{\theta}_j||^2 \leq R_j^2\\ 
0, & \text{otherwise} \ \ 
\end{matrix}\right. 
\end{equation} 
Note that the expressions for $u^*_{ij}$ given by eqs. (\ref{globJ}) and (\ref{globJ1}) are equivalent and will be used interchangeably in the subsequent analysis.

\subsection{The SPCM algorithm}
\label{subsecSPCM}
Taking into account the previous short description of its main features, the SPCM algorithm is summarized as follows.

\begin{algorithm}[htpb!]
\caption{ [$\Theta$, $\Gamma$, $U$] = SPCM($X$, $m$)} 
\label{alg:spcm}
\begin{algorithmic}[1]
\doublespacing
\Require {$X$, $m$}  
\State $t=0$
\Statex {\Comment{Initialization of $\boldsymbol{\theta}_j$'s part}}
\State \textbf{Initialize:} $\boldsymbol{\theta}_j(t)$ {\it via FCM algorithm}
\Statex {\Comment{Initialization of $\gamma_j$'s part}}
   \State \textbf{Set:} $\gamma_j=\frac{ \sum_{i=1}^N u^{FCM}_{ij} \|\mathbf{x}_i-\boldsymbol{\theta}_j(t)\|^2}{ \sum_{i=1}^N u^{FCM}_{ij} }$, $j=1,...,m$ 
	\State \textbf{Set:} $\lambda=K\frac{\bar{\gamma}}{p(1-p)\mathrm{e}^{2-p}}$, where $\bar{\gamma}=\min\limits_{j=1,\ldots,m}\gamma_j$
\Repeat 
\Statex {\Comment {Update $U$ part}}
\State \textbf{Update} $U(t)$ {\it via eq.~\eqref{globJ}, as described in the text}
\Statex {\Comment{Update $\Theta$ part}}
\State \text{$\boldsymbol{\theta}_j(t+1)=\left.{\sum\limits_{i=1}^N u_{ij}(t)\mathbf{x}_i} \middle/ {\sum\limits_{i=1}^N u_{ij}(t)} \right.$, $j=1,...,m$} 
\State $t=t+1$ 
%
\Until{the change in $\boldsymbol{\theta}_j$'s between two successive iterations becomes sufficiently small}\\ 
\Return {$\Theta$, $\Gamma=\{\gamma_1,\ldots,\gamma_m\}$, $U$}
\end{algorithmic}
\end{algorithm}
It is noted that after the termination of the algorithm an additional step is required, in order to identify and remove possibly duplicated clusters. 

The worst case computational complexity of (the main body of) SPCM is $O((\epsilon+2)Nm\cdot iter)$, where $\epsilon$ is the number of iterations in the bisection method (which have very light computational complexity\footnote{In our case $\epsilon$ is fixed to 30, which implies an accuracy of $10^{-10}$.}) and $iter$ is the number of iterations performed by the algorithm. Note, however, that the actual complexity is much less since at each iteration the bisection method is activated only for a small fraction of $u_{ij}$'s. As it is shown experimentally in \cite{Xena16} the computational complexity of SPCM is slightly increased compared to that of PCM. This is the price to pay for the better quality results of SPCM compared to PCM.

\section{Convergence proof of the SPCM}

In the sequel, a proof of the convergence of the SPCM is provided. Note that, in principle, the proof holds for any choice of (fixed) $\gamma_j$'s, not only for the one given in eq. (\ref{initgamma}).

Before we proceed, we note that the cost function associated with SPCM (eq. (\ref{Jspcm})) can be recasted as
\begin{equation}
J_{SPCM}(U,\Theta)=\sum_{j=1}^m J_j(\mathbf{u}_j,\boldsymbol{\theta}_j) \ \equiv \sum_{j=1}^m \left[ \sum_{i=1}^N \overbrace{u_{ij} \|\mathbf{x}_i-\boldsymbol{\theta}_j\|^2 + \gamma_j \sum_{i=1}^N (u_{ij} \ln u_{ij} - u_{ij}) + \lambda u_{ij}^p}^{h(u_{ij},\boldsymbol{\theta}_j)} \right]
\label{Jspcm1}
\end{equation}
where $\mathbf{u}_j=[u_{1j},\ldots,u_{Nj}]^T$. Since (a) $u_{ij}$'s, $j=1,\ldots,m$, are not interrelated to each other, for a specific $\mathbf{x}_i$, (b) $u_{ij}$'s, $i=1,\ldots,N$ are related exclusively with $\boldsymbol{\theta}_j$ and vice versa and (c) $\boldsymbol{\theta}_j$'s are not interrelated to each other, minimization of $J_{SPCM}(U,\Theta)$ can be considered as the minimization of $m$ independent cost functions $J_j$'s, $j=1,\ldots,m$. Thus, in the sequel, we focus on the minimization of a specific $J_j(\mathbf{u}_j,\boldsymbol{\theta}_j)$ and, for the ease of notation, we drop the index $j$, i.e., when we write $J(\mathbf{u},\boldsymbol{\theta})$, $\mathbf{u}=[u_1,\ldots,u_N]^T$, we refer to a $J_j(\mathbf{u}_j,\boldsymbol{\theta}_j)$.

The proof is given under the very mild assumption that for each one cluster at least one equation $f(u_i)=0$, $i=1,\ldots,N$ has two solutions at each iteration of SPCM ({\it Assumption 1}). This is a rational assumption, since if this does not hold at a certain iteration, the algorithm cannot identify new locations for $\boldsymbol{\theta}$ at the next iteration. In subsection \ref{subsecKpara}, it is shown how this assumption can always be fulfilled.

Some definitions are now in order. Let ${\cal M}$ be the set containing all the $N \times 1$ vectors $\mathbf{u}$ whose elements lie in the union $\{0\} \cup [u^{min},\ u^{\max}]$, i.e. ${\cal M}=\left(\{0\}\cup[u^{min},u^{max}]\right)^N$.
 Also, let ${\cal R}^{l}$ be the space where the vector $\boldsymbol{\theta}$ lives. The SPCM algorithm produces a sequence $(\mathbf{u}^{(t)},\boldsymbol{\theta}^{(t)})\arrowvert_{t=0}^{\infty}$, which will be examined in terms of its convergence properties.

Let 
$$G: {\cal M} \rightarrow {\cal R}^{l},\ \text{with} \ G(\mathbf{u})=\boldsymbol{\theta}$$
where $G$ is calculated via the following equation

\begin{equation}
\label{eq2}
{\mbox{\boldmath $\theta$}} = \frac{\sum_{i=1}^N u_i {\mathbf{x}}_i }{\sum_{i=1}^N u_i}
\end{equation}

and
$$F:{\cal R}^{l} \rightarrow {\cal M},\  \text{with} \ F(\boldsymbol{\theta})=\mathbf{u}$$
where $F$ is calculated via eq. (\ref{globJ1}).
%
Then, the SPCM operator 
$T: {\cal M} \times {\cal R}^{l} \rightarrow {\cal M} \times {\cal R}^{l}$ is defined as
\begin{equation}
\label{T}
T=T_2 \circ T_1
\end{equation}
where 
\begin{equation}
\label{T1}
T_1: {\cal M} \times {\cal R}^{l} \rightarrow {\cal M},\ \ T_1(\mathbf{u},\boldsymbol{\theta})=F(\boldsymbol{\theta})
\end{equation}
and 
\begin{equation}
\label{T2}
T_2: {\cal M} \rightarrow {\cal M} \times {\cal R}^{l},\ \ T_2(\mathbf{u})=(\mathbf{u},G(\mathbf{u}))
\end{equation}
For operator $T$ we have that
$$T(\mathbf{u},\boldsymbol{\theta})=(T_2 \circ T_1)(\mathbf{u},\boldsymbol{\theta}) = T_2(T_1(\mathbf{u},\boldsymbol{\theta}))=T_2(F(\boldsymbol{\theta})) =$$ 
$$(F(\boldsymbol{\theta}),G(F(\boldsymbol{\theta}))) = (F(\boldsymbol{\theta}),(G \circ F)(\boldsymbol{\theta}))$$
Thus, the iteration of SPCM can be expressed in terms of $T$ as
$$(\mathbf{u}^{(t)},\boldsymbol{\theta}^{(t)}) = T(\mathbf{u}^{(t-1)},\boldsymbol{\theta}^{(t-1)}) = (F(\boldsymbol{\theta}^{(t-1)}),(G \circ F)(\boldsymbol{\theta}^{(t-1)}))$$

The above decomposition of $T$ to $T_1$ and $T_2$ will facilitate the subsequent convergence analysis, since certain properties for $T$ can be proved relying on $T_1$ and $T_2$ (and, ultimately, on $F$ and $G$).

{\em Remark 1:} Note that $F$ (and as a consequence $T_1$) are, in general, not continuous (actually they are piecewise continuous).

In the sequel some required definitions are given. Let $Z: X \rightarrow X$ ($X \subset {\cal R}^p$)  be a point-to-point map that gives rise to an iterative algorithm $z(t)=Z(z(t-1))$, which generates a sequence $z(t)|_{t=0}^{\infty}$, for a given $z(0)$. 
A {\em fixed point} $z^*$ of $Z$ is a point for which $Z(z^*)=z^*$. Also, we say that $Z$ is {\em strictly monotonic} with respect to a (continuous) function $g$ if $g(Z(z))<g(z)$, whenever $z$ is not a fixed point of $Z$.
Having said the above, we can now state the following theorem that will be proved useful in the sequel:



{\em Theorem 1 \cite{Meye76} \footnote{This is a direct combination of Theorem 3.1 and Corollary 3.2 in \cite{Meye76}.} :} Let $Z: X \rightarrow X$ ($X \in {\cal R}^p$) be a point-to-point map that gives rise to an iterative algorithm $z(t)=Z(z(t-1))$, which generates a sequence $z(t)|_{t=0}^{\infty}$, for a given $z(0)$. Supposing that:

(i) $Z$ is strictly monotonic with respect to a continuous function $g:X \rightarrow {\cal R}$,

(ii) $Z$ is continuous on $X$,

(iii) the set of all points $z(t)|_{t=0}^{\infty}$ is bounded and

(iv) the number of fixed points having any given value of $g$ is finite

then 

the algorithm corresponding to $Z$ will converge to a fixed point of $Z$ regardless where it is initialized in $X$ \footnote{Actually, this theorem has been stated for the more general case where $Z$ is a one-to-many mapping \cite{Meye76}. The present form of the theorem is for the special case where $Z$ is a one-to-one mapping, which is the case for SPCM.}.

{\vskip 15pt}

In the SPCM case, $Z$ is the mapping $T$ (SPCM operator) defined by eq. (\ref{T}) and $g$ is the cost function $J$. Due to the fact that SPCM has been resulted from the minimization of $J$, it turns out that its fixed points $(\mathbf{u}^*,\boldsymbol{\theta}^*)$ satisfy $\nabla J|_{(\mathbf{u},\boldsymbol{\theta}) }=\boldsymbol{0}$.

Although the general strategy to prove convergence for an algorithm is to show that it fulfills the requirements of the convergence theorem, this cannot be adopted in this straightforward manner in this framework. The reason is that Theorem 1 requires continuity of $T$, which is not guaranteed in the SPCM case due to $T_2$ ($F$) (see eq. (\ref{globJ1})), which is not continuous in its domain (which is the convex hull of $X$, $CH(X)$)\footnote{Due to its updating (eq. (\ref{theta})), $\boldsymbol{\theta}$ will always lie in $CH(X)$, provided that its initial position lies in at least one hypersphere of radius $R$ centered at a data point.}. However, it is continuous on certain subsets of $CH(X)$. This fact will allow the use of Theorem 1 for certain small regions where continuity is preserved.

\begin{figure}[htpb!]
\centering
\subfloat[]{\includegraphics[width=0.48\textwidth]{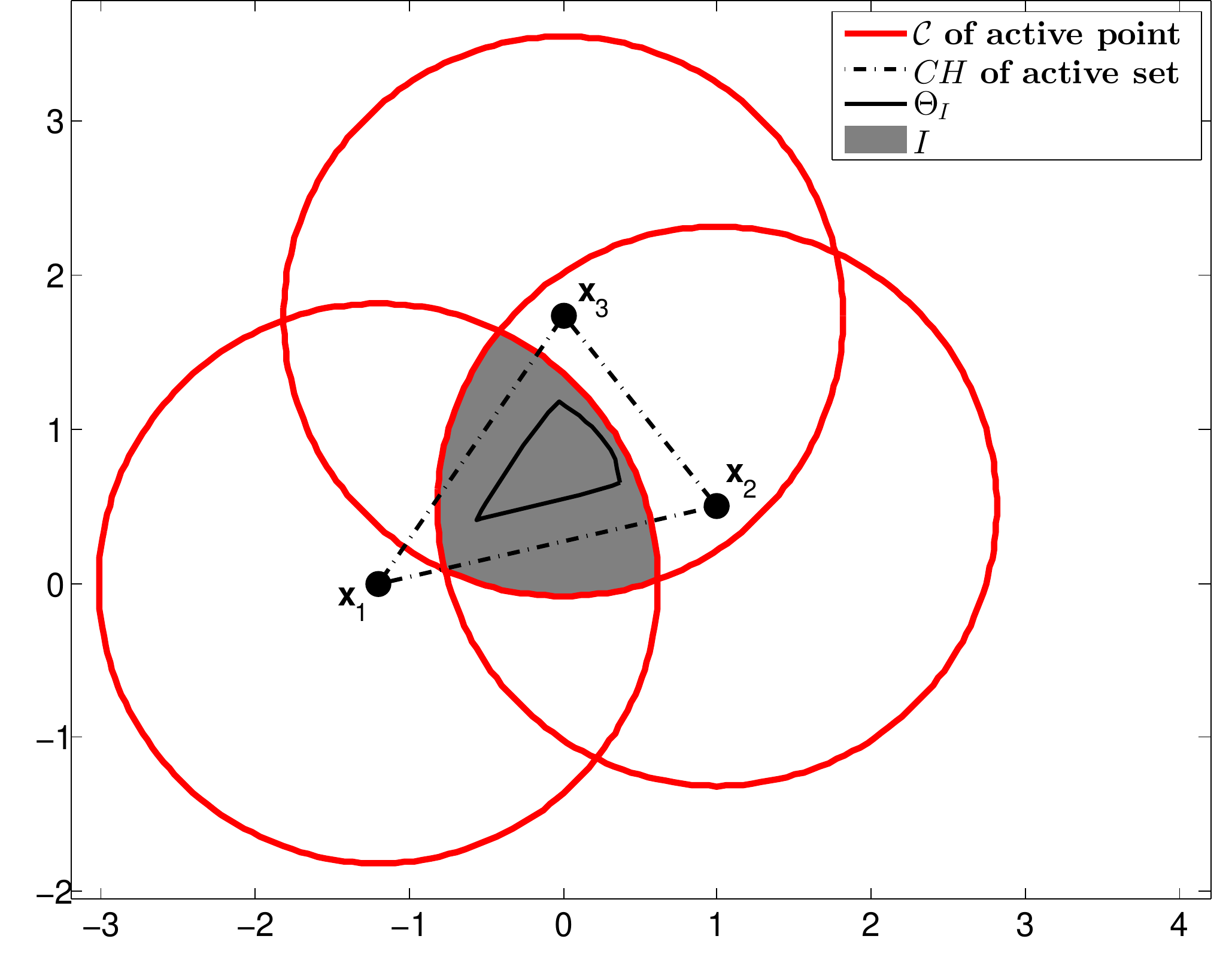}\label{figure1a}}
\hfil
\centering
\subfloat[]{\includegraphics[width=0.49\textwidth]{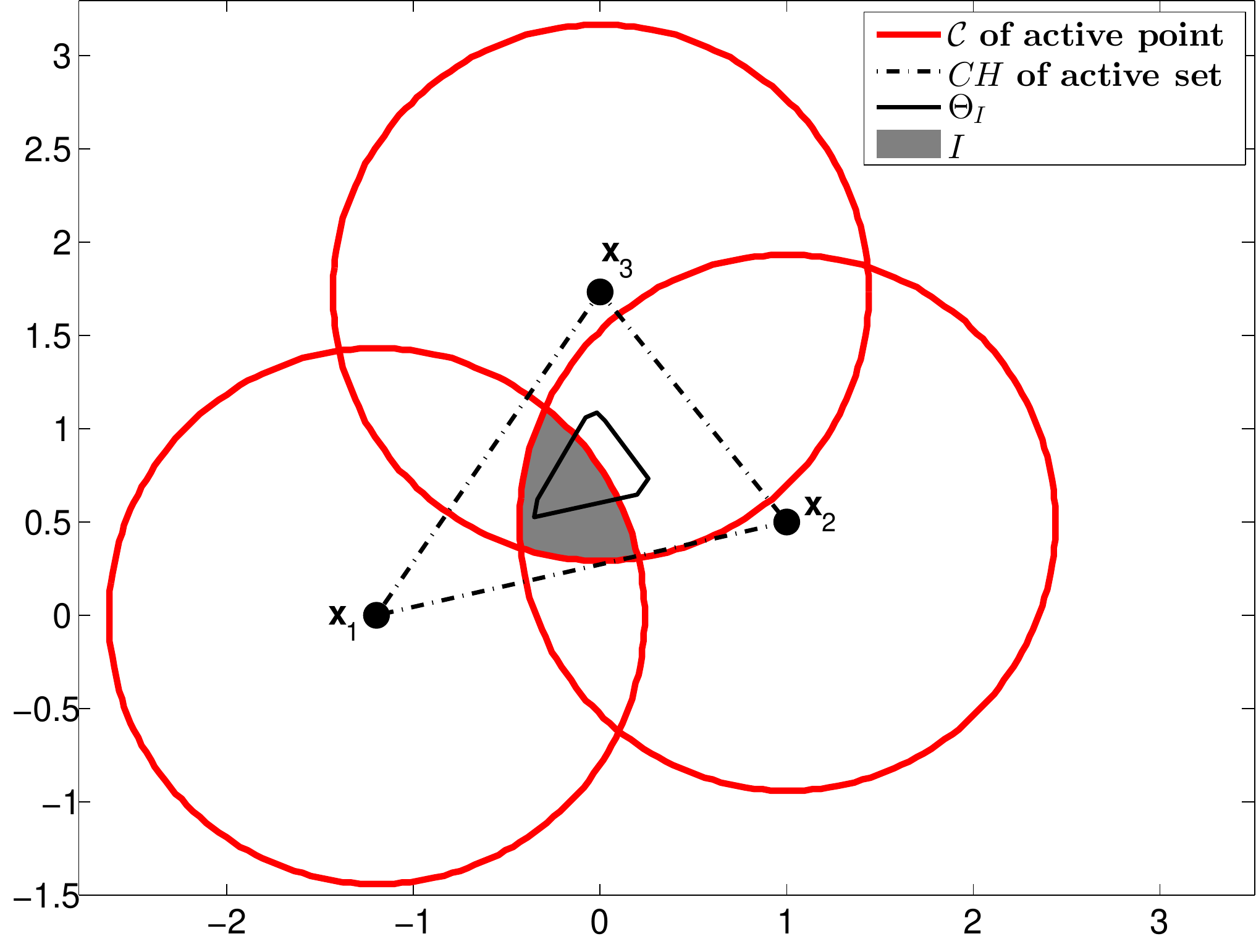}\label{figure1b}}
\hfil
\centering{\caption{An active set of $k=3$ points in cases when (a) $\Theta_I\subset I$ and (b) $\Theta_I\not\subset I$ }\label{defi1}}
\end{figure}

Some additional definitions are now in order. Without loss of generality, let $I= ( \cap_{i=1}^k int({\cal C}_i) )$; that is $I$ is the (nonempty) intersection of the interiors of the hyperspheres of radius $R$ (eq. (\ref{R2})) that correspond to $\mathbf{x}_i$'s, $i=1,\ldots,k$ (see Fig.~\ref{defi1})\footnote{Clearly, by reordering the data points we can take all the possible corresponding $I$ intersections.}. Note that for $\boldsymbol{\theta} \in I$ the above $k$ points will have $u_i>0$. The set of all data points that have $u_i>0$ form the so-called {\em active set}, while the points themselves are called {\em active points}. In addition, an active set $X_q$ is called {\em valid} if its corresponding intersection of hyperspheres $I_q$ is nonempty. Finally, the points with $u_{i}=0$ are called {\it inactive}.

Let also 
\begin{equation}
\label{UI}
U_I= \{ \mathbf{u}=[u_1,\ldots,u_k]: \mathbf{u} = F(\boldsymbol{\theta}),\ \text{for}\ \boldsymbol{\theta} \in I \}
\end{equation}
be the set containing all possible values of the degrees of compatibility, $u_i$, of $\boldsymbol{\theta}$ with the $k$ active $\mathbf{x}_i$'s. Clearly, $u_i$'s are computed via the first branch of eq. (\ref{globJ1}) and $F$ is continuous in this specific case (as it will be explicitly shown later). Also, let
\begin{equation}
\label{thetaI}
\Theta_I= \{ \boldsymbol{\theta}: \boldsymbol{\theta} = G(\mathbf{u}),\ \text{for}\ \mathbf{u} \in U_I \}
\end{equation}
(see Fig.~\ref{defi1} for the possible scenarios for $\Theta_I$).
Three observations are now in order: 
\begin{itemize}
\item First, due to the fact that $u_i$'s are independent from each other, $U_I$ can also be expressed as 
\begin{equation}
\label{UI1}
U_I= \Pi_{i=1}^k [u^{min}, u_i^{max}] 
\end{equation}
where $\Pi$ denotes the Cartesian product and $u_i^{max}$ is the maximum possible value $u_i$ can take, provided that $\boldsymbol{\theta} \in I$ (clearly $u_i^{max} \leq u^{max}$). 

\item If at a certain iteration $t$ of SPCM, $\boldsymbol{\theta}(t) \in I$, $\Theta_I$ contains all possible positions of $\boldsymbol{\theta}(t+1)$.

\item $\Theta_I$ always lies in the convex hull of the associated active set.
\end{itemize}

In the sequel, we proceed by showing the following facts, that are preliminary for the establishment of the final convergence result. Specifically, we will show that 
\begin{itemize}
\item (A) $J(\mathbf{u},\boldsymbol{\theta})$ decreases at each iteration of the SPCM operator $T$
\item (B) $T$ is continuous on every region $U_I \times I$ that corresponds to a valid active set.
\item (C) The sequence produced by the algorithm is bounded
\item (D) The fixed points corresponding to a certain valid active set (if they exist) are strict local minima of $J$ and they are finite.
\end{itemize}

\subsubsection{Proof of item (A)}

%
%

To achieve this goal, we prove first the following two lemmas

{\vskip 10pt}
{\em Lemma 1:} Let $\phi: {\cal M} \rightarrow {\cal R}$, $\phi(\mathbf{u})=J(\mathbf{u},\boldsymbol{\theta})$, where $\boldsymbol{\theta}$ is fixed. Then $\mathbf{u}^*$ is the global minimum solution of $\phi$ if and only if $\mathbf{u}^*=F(\boldsymbol{\theta})$, where $F$ is defined as in eq.~(\ref{globJ}).
{\vskip 5pt}
{\em Proof:} We proceed by showing that 

\noindent (a) the unique point $\mathbf{u}^*$ that satisfies the KKT conditions for the minimization problem
\begin{equation}
\label{min-prob}
\begin{array}{l}
\min \phi(\mathbf{u}) \\
\text{subject\ to\ }\ \ u_i \geq 0,\ \ \ \ \ \ \ i=1,\ldots,N \\
\text{and}\ \ \ \ \ \ \ \ \ \ 1-u_i \geq 0,\ i=1,\ldots,N
\end{array} 
\end{equation}
is the one determined by eq.~(\ref{globJ}) and 

\noindent (b) this point is a minimizer of $J$, which implies (due to the uniqueness) that it is the global minimizer.

Let $\mathbf{u}^*=[u_i^*]$ be a point that satisfies the KKT conditions for (\ref{min-prob}). Then we have
\begin{equation}
\label{constr1}
(i)\ u^*_i \geq 0,\ \ \ (ii)\ 1-u^*_i \geq 0
\end{equation}

\begin{equation}
\label{constr2}
(i)\ \exists\ \kappa_i \geq 0:\ \kappa_i u^*_i=0,\ \ \ \ (ii)\ \exists\ \tau_i \geq 0:\ \tau_i (1-u^*_i)=0
\end{equation}
and
\begin{equation}
\label{constr3}
\frac{\partial {\cal L} (\mathbf{u})}{\partial u_i}|_{\mathbf{u}=\mathbf{u}^*} = 0
\end{equation}
where ${\cal L}(\mathbf{u})$ is the Lagrangian function defined as
\begin{equation}
\label{lagra}
{\cal L} (\mathbf{u}) = \phi(\mathbf{u}) - \sum_{i=1}^N \kappa_i u_i - \sum_{i=1}^N \tau_i (1-u_i)
\end{equation}

Recalling eq.~\eqref{Jspcm}, $\phi(\mathbf{u})$ can be written as
\begin{equation}
\label{phi}
\phi(\mathbf{u})= \sum_{i=1}^N \overbrace{[u_i ||{\mathbf{x}}_i-{\mbox{\boldmath $\theta$}}||^2 + \gamma (u_i \ln u_i -u_i) + \lambda u_i^p]}^{h(u_i;{\mbox{\boldmath $\theta$}})}
\end{equation}
where $h(u_i;{\mbox{\boldmath $\theta$}})$ is a function of $u_i$ for a fixed value of ${\mbox{\boldmath $\theta$}}$. Noting that all $u_i$'s are computed independently from each other, for fixed $\boldsymbol{\theta}$, it is easy to verify that, for a specific $u_i$ it is 
$$\frac{\partial \phi(\mathbf{u})}{\partial u_i} = \frac{\partial h(u_i;{\mbox{\boldmath $\theta$}}) }{\partial u_i} = ||{\mathbf{x}}_i-{\mbox{\boldmath $\theta$}}||^2 + \gamma \ln u_i + \lambda p u_i^{p-1} \equiv f(u_i)$$

As a consequence, eq.~(\ref{constr3}) gives
\begin{equation}
\label{constr3-1}
||{\mathbf{x}}_i-{\mbox{\boldmath $\theta$}} ||^2 + \gamma \ln u_i^* + \lambda p u_i^{*p-1} - \kappa_i +\tau_i=0
\end{equation}
We will prove next that $\kappa_i=0$ and $\tau_i=0$, for $i=1,\ldots,N$; that is, the constraints on $u_i$'s are inactive, i.e., the optimum of $\phi(\mathbf{u})$ lies always in the region defined by the constraints. Assume, on the contrary, that there exists $\kappa_s>0$. From eq.~(\ref{constr2}-(i)) it follows that $u_s^*=0$ and from eq.~(\ref{constr2}-(ii)) that $\tau_s=0$. Taking into account that $\lim_{ u_s^*\rightarrow 0^+} \left(\gamma \ln u_s^* + \lambda p u_s^{*\ p-1}\right) = +\infty$ \footnote{Utilization of the L' Hospital rule gives that $\lim_{x \rightarrow 0^+} x^{1-p} \ln x = 0$ ($p<1$). Then $\lim_{x \rightarrow 0^+} (\ln x + \beta \frac{1}{x^{1-p}}) = \lim_{x \rightarrow 0^+} \frac{x^{1-p} \ln x + \beta}{x^{1-p}}=+\infty$, for $\beta>0$. Setting $x=u_s^*$, $\beta=\frac{\lambda p}{\gamma}$, the claim follows.} and applying eq.~(\ref{constr3-1}) for $u_s^*$ we have
\begin{equation}
\label{contradi-k}
||{\mathbf{x}}_s-{\mbox{\boldmath $\theta$}} ||^2 + \infty  = \kappa_s\ \ \text{or} \ \ \kappa_s=+\infty
\end{equation}
which contradicts the fact that $\kappa_s$ is finite.

Assume next that there exists $\tau_s>0$. From eq.~(\ref{constr2}-(ii)) it follows that $u_s^*=1$ and from eq.~(\ref{constr2}-(i)), it is $\kappa_s=0$. Applying eq.~(\ref{constr3-1}) for $u_s^{*}$ and substituting the above we have
\begin{equation}
\label{contradi-t}
||{\mathbf{x}}_s-{\mbox{\boldmath $\theta$}} ||^2 + \gamma \ln 1  + \lambda p 1^{p-1} + \tau_s=0\ \ \text{or} \ \ \tau_s=-||{\mathbf{x}}_s-{\mbox{\boldmath $\theta$}} ||^2 - \lambda p < 0
\end{equation}
which contradicts the fact that $\tau_s>0$. Thus $\tau_s=0$.

Since $\kappa_i=\tau_i=0$, for all $i$, eq.~(\ref{constr3-1}) becomes
\begin{equation}
\label{u-ij}
||{\mathbf{x}}_i-{\mbox{\boldmath $\theta$}} ||^2 + \gamma \ln u_i^* + \lambda p u_i^{*\ p-1} \equiv f(u_i^{*})= 0,\ \ i=1,\ldots,N
\end{equation}
Note that the algorithm relies on eq.~\eqref{u-ij} in order to derive the updating formula of eq.~\eqref{globJ} (thus step (a) has been shown). We proceed now to show that the point corresponding to eq.~\eqref{globJ} (derived through eq.~\eqref{u-ij}) minimizes $J$. We consider the following two cases: 

$\bullet$ $u_i^{*}$ is given by the first branch of eq. (\eqref{globJ}). This implies that
 $f(u_i)=0$ has two solutions $u_i^{\{1\}}$ and $u_i^{\{2\}}$ ($u_i^{\{1\}}<u_i^{\{2\}}$) and 
$u_i^{\{2\}}>\left(\frac{\lambda (1-p)}{\gamma_j}\right)^{\frac{1}{1-p}} (=u^{min})$ (figures 1a, 1d).
%
%
Taking into account the definition of $h(u_i;{\mbox{\boldmath $\theta$}})$ in eq.~\eqref{phi}, it can be shown (Proposition 5, \cite{Xena16}) that the maximum of the two solutions $u_i^{\{1\}}$, $u_i^{\{2\}}$ ($u_i^{\{1\}}<u_i^{\{2\}}$) is the one that minimizes $h(u_i;{\mbox{\boldmath $\theta$}})$ and, as a consequence, $\phi(\mathbf{u})$ also (which equals to $J(\mathbf{u},\boldsymbol{\theta})$) with $\boldsymbol{\theta}$ fixed.

$\bullet$ $u_i^{*}$ is given by the second branch of eq. (\ref{globJ}). In this case we have that either (i) $f(u_i)$ is strictly positive, which implies that $J(\mathbf{u},\boldsymbol{\theta})$ is strictly increasing with respect to $u_i$ (case shown in figures 1b, 1e) or (ii) $h(u_i^{\{2\}},\boldsymbol{\theta}) \geq h(0,\boldsymbol{\theta})=0$ (case shown in figures 1c, 1f). In both (i) and (ii) cases, $J(\mathbf{u},\boldsymbol{\theta})$ is minimized with respect to $u_i$ only for $u_i=0$ (the second branch of eq.~\eqref{globJ}).

From the above, it follows that $\mathbf{u}^*$ is the global minimum solution of $\phi$ if and only if $\mathbf{u}^*$ is given by eq.~(\ref{globJ}).  Q.E.D.

{\vskip 10pt}
{\em Lemma 2:} Let $\psi:{\cal R}^{l} \rightarrow {\cal R}$, with $\psi(\boldsymbol{\theta})=J(\mathbf{u},\boldsymbol{\theta})$, with $\mathbf{u} \in U_I$ being fixed. Then, $\boldsymbol{\theta}^*$ ($\in\Theta_I$) is the unique global minimum of $\psi$ if and only if $\boldsymbol{\theta}^*=G(\mathbf{u})$, where $G$ is calculated as in 
eq. (\ref{eq2}).
{\vskip 5pt}
{\em Proof:} In contrast to the situation in Lemma 1, the minimization of $\psi(\boldsymbol{\theta})$ with respect to $\boldsymbol{\theta}$ is an unconstrained optimization problem. The stationary points of $\psi(\boldsymbol{\theta})$ are obtained as the solutions of the equations
\begin{equation}
\label{grad-psi}
\frac{\partial \psi}{\partial {\mbox{\boldmath $\theta$}}} = \frac{\partial}{\partial {\mbox{\boldmath $\theta$}}} \left [ \sum_{i=1}^N \left ( u_i ||{\mathbf{x}}_i-{\mbox{\boldmath $\theta$}}||^2 + \gamma (u_i\ln u_i - u_i) + \lambda u_i^p \right ) \right ] = 2 \sum_{i=1}^N u_i ({\mbox{\boldmath $\theta$}}-{\mathbf{x}}_i) = \mathbf{0},
\end{equation}
which, after some manipulations, give
\begin{equation}
\label{theta*}
{\mbox{\boldmath $\theta$}}^* = \frac{\sum_{i=1}^N u_i {\mathbf{x}}_i }{\sum_{i=1}^N u_i }.
\end{equation}
Also, it is 
\begin{equation}
H_{\psi} \equiv \frac{\partial^2 \psi}{\partial {\mbox{\boldmath $\theta$}}^2} = \overbrace{2\sum_{i=1}^N u_i }^{b} I^l
\label{eikosi}
\end{equation}
where $I^l$ is the $l \times l$ identity matrix. Under {\it Assumption 1}, stating that at least one $u_i$ is computed by the first branch of eq. (\ref{globJ}), it is $b>0$. Therefore, $\psi$ is a convex function over ${\cal R}^l$, with a unique stationary point, given by eq. (\ref{theta*}), which is the unique global minimum of $\psi(\boldsymbol{\theta})$. Q.E.D.

{\vskip 7pt}
Combining now the previous two lemmas, we are in a position to prove the following lemma.
{\vskip 10pt}
{\em Lemma 3:} Consider a valid active set, whose corresponding hyperspheres intersection is denoted by $I$. Let
$$S=\{ (\mathbf{u},\boldsymbol{\theta})=([u_1,\ldots,u_k],\boldsymbol{\theta}) \in U_I \times I:
 \nabla J|_{(\mathbf{u},\boldsymbol{\theta}) }=\boldsymbol{0} \ \text{with} \ u_i\ \text{being} \ \text{the}$$
\begin{equation}
\label{S}
\text{largest of the two solutions of } f_{\boldsymbol{\theta}}(u_i)=0,\ i=1,\ldots,k\}\ \footnote{In the sequel, we insert $\boldsymbol{\theta}$ as subscript in the notation of $f$ in order to show explicitly the dependence of $u_i$ from $\boldsymbol{\theta}$.}
\end{equation}
Then $J$ is continuous over $U_I \times I$ and 
$$J(T(\mathbf{u},\boldsymbol{\theta})) < J(\mathbf{u},\boldsymbol{\theta}),\ \text{if}\ (\mathbf{u},\boldsymbol{\theta}) \notin S$$
{\vskip 5pt}
{\em Proof:} 
Since $\{y \rightarrow ||y||^2\}$, $\{y \rightarrow \ln y\}$, $\{ y \rightarrow y^p \}$ are continuous and $J$ is a sum of products of such functions, it follows that $J$ is continuous on $U_I \times I$. 
Let $(\mathbf{u},\boldsymbol{\theta}) \notin S$. Recalling that 
$$T(\mathbf{u},\boldsymbol{\theta})=(F(\boldsymbol{\theta}),(G \circ F)(\boldsymbol{\theta})) = (F(\boldsymbol{\theta}),G(F(\boldsymbol{\theta})))$$
we have
\begin{equation}
\label{lem3-1}
J(T(\mathbf{u},\boldsymbol{\theta})) = J( (F(\boldsymbol{\theta}),G(F(\boldsymbol{\theta}))) )      
\end{equation}
Applying Lemma 1 for fixed $\boldsymbol{\theta}$, we have that $F(\boldsymbol{\theta})$ is the unique global minimizer of $J$. Thus, 
\begin{equation}
\label{lem3-3}
J(F(\boldsymbol{\theta}),\boldsymbol{\theta}) < J(\mathbf{u},\boldsymbol{\theta})  
\end{equation}
Applying Lemma 2 for fixed $F(\boldsymbol{\theta})$, we have that $G(F(\boldsymbol{\theta}))$ is the unique global minimizer of $J$. Thus, it is 
\begin{equation}
\label{lem3-2}
J(F(\boldsymbol{\theta}),G(F(\boldsymbol{\theta}))) < J(F(\boldsymbol{\theta}),\boldsymbol{\theta})
\end{equation}

From eqs. (\ref{lem3-1}), (\ref{lem3-3}) and (\ref{lem3-2}), it follows that
$$J(T(\mathbf{u},\boldsymbol{\theta})) < J(\mathbf{u},\boldsymbol{\theta}),\ \ \text{for}\ (\mathbf{u},\boldsymbol{\theta}) \notin S$$ Q.E.D.

{\vskip 7pt}
{\em Remark 2:} It is noted that although the above proof has been focused on the $k$ (active) points, its generalization that takes also into account the rest data points is straightforward since $u_i=0$, for $i=k+1,\ldots,N$ and the corresponding terms $h(u_i,\boldsymbol{\theta})$ that contribute to $J$ are $0$.

{\vskip 7pt}
{\em Remark 3:} Taking into account that SPCM has been resulted from the minimization of $J$ ($\nabla J|_{(\mathbf{u},\boldsymbol{\theta})}=\boldsymbol{0}$) on a $U_I\times I$ corresponding to an active set, it follows that $S$ contains all the fixed points of $T$, which (as will be shown later) are local minima of the cost function $J$ (of course, $J$ may have additional local minima than those belong to $S$ which are not accessible by the algorithm).

Now we proceed by showing that $T$ decreases $J$, in the whole domain $(\{0\} \cup [u^{min},u^{max}])^N \times CH(X)$.

{\vskip 10pt}
{\em Lemma 4:} The strict monotonically decreasing property of $T$ with respect to $J$ remains valid in the domain $(\{0\} \cup [u^{min},u^{max}])^N \times CH(X)$ excluding the fixed points of $T$ of each valid active set.
{\vskip 5pt}
{\em Proof:} Let $(\bar{\mathbf{u}},\bar{\boldsymbol{\theta}})$ be the outcome of SPCM at a specific iteration, $\hat{\mathbf{u}}=F(\bar{\boldsymbol{\theta}})$ be the $\mathbf{u}$ for the next iteration and $\hat{\boldsymbol{\theta}}=G(\hat{\mathbf{u}})$ be the subsequent $\boldsymbol{\theta}$. Recall that the ordering of the updating is
\begin{equation}
\label{order}
\bar{\mathbf{u}} \rightarrow \bar{\boldsymbol{\theta}} \rightarrow \hat{\mathbf{u}} \rightarrow \hat{\boldsymbol{\theta}}
\end{equation}
 We define 
$$\bar{\Gamma} =\{ i\ :\ \bar{u}_i\ \text{is\ computed\ via\ the\ second\ branch\ of\ eq.}~(\ref{globJ}) \}$$ 
and
$$\hat{\Gamma} =\{ i\ :\ \hat{u}_i\ \text{is\ computed\ via\ the\ second\ branch\ of\ eq.}~(\ref{globJ}) \}$$ 

Recalling that $h(u_i;{\mbox{\boldmath $\theta$}}) = u_i ||{\mathbf{x}}_i - {\mbox{\boldmath $\theta$}}||^2 + \gamma (u_i \ln u_i - u_i) + \lambda u_i^p$, we can write
\begin{equation}
\label{Jbar}
J(\bar{\mathbf{u}},\bar{\boldsymbol{\theta}}) = \overbrace{\sum_{i \in \bar{\Gamma} \cap \hat{\Gamma} } h(\bar{u}_i;\bar{ {\mbox{\boldmath $\theta$}} }) }^{\bar{A}_1} + \overbrace{\sum_{i \in \ \tilde{ }\ \bar{\Gamma}  \cap \hat{\Gamma} } h(\bar{u}_i;\bar{ {\mbox{\boldmath $\theta$}} }) }^{\bar{A}_2} + \overbrace{\sum_{i \in \ \tilde{ }\ \hat{\Gamma} } h(\bar{u}_i;\bar{ {\mbox{\boldmath $\theta$}} }) }^{\bar{A}_3}
\end{equation}
and
\begin{equation}
\label{Jbar2}
J(\hat{\mathbf{u}},\bar{\boldsymbol{\theta}}) = \overbrace{\sum_{i \in \bar{\Gamma} \cap \hat{\Gamma} } h(\hat{u}_i;\bar{ {\mbox{\boldmath $\theta$}} }) }^{\hat{A}_1} + \overbrace{\sum_{i \in \ \tilde{ }\ \bar{\Gamma}  \cap \hat{\Gamma} } h(\hat{u}_i;\bar{ {\mbox{\boldmath $\theta$}} }) }^{\hat{A}_2} + \overbrace{\sum_{i \in \ \tilde{ }\ \hat{\Gamma} } h(\hat{u}_i;\bar{ {\mbox{\boldmath $\theta$}} }) }^{\hat{A}_3}
\end{equation}
where $\tilde{ }\ \Gamma$ denotes the complement of $\Gamma$.

Focusing on $\bar{A}_1$ and $\hat{A}_1$, we have that $h(\bar{u}_i;\bar{ {\mbox{\boldmath $\theta$}} }) = h(\hat{u}_i;\bar{ {\mbox{\boldmath $\theta$}} }) = 0$, since $i \in \bar{\Gamma} \cap \hat{\Gamma}$. Thus 
\begin{equation}
\label{A1}
\hat{A}_1 = \bar{A}_1=0
\end{equation}

Considering $\bar{A}_2$ and $\hat{A}_2$, since $i \in \hat{\Gamma}$, we have $\hat{u}_i=0$. 
Thus, taking into account the order of updating (eq. (\ref{order})) and Lemma 1, we have $(0 =) \ h(\hat{u}_i;\bar{ {\mbox{\boldmath $\theta$}} } ) < h(\bar{u}_i;\bar{ {\mbox{\boldmath $\theta$}} } )$. Thus, it follows that
\begin{equation}
\label{A2}
\hat{A}_2 < \bar{A}_2
\end{equation}

Finally, focusing on $\bar{A}_3$ and $\hat{A}_3$, since $i \in \ \tilde{ }\ \hat{\Gamma}$, the argumentation of Lemma 1 implies that the global minimum of $h(u_i;\bar{ {\mbox{\boldmath $\theta$}} } )$ is met at $\hat{u}_i=u_i^{\{2\}}$. Thus, taking also into account the order of updating in eq. (\ref{order}), it is $h(\hat{u}_i;\bar{ {\mbox{\boldmath $\theta$}} } ) < h(\bar{u}_i;\bar{ {\mbox{\boldmath $\theta$}} } )$. Therefore, it is
\begin{equation}
\label{A3}
\hat{A}_3 < \bar{A}_3
\end{equation}

Combining eqs.~(\ref{A1}), (\ref{A2}) and (\ref{A3}) it follows that
\begin{equation}
\label{assu2}
J(\hat{\mathbf{u}},\bar{\boldsymbol{\theta}}) < J(\bar{\mathbf{u}},\bar{\boldsymbol{\theta}})
\end{equation}
Also, lemma 2 gives
\begin{equation}
\label{assu3}
J(\hat{\mathbf{u}},\hat{\boldsymbol{\theta}}) < J(\hat{\mathbf{u}},\bar{\boldsymbol{\theta}})\ \  \footnote{Considering a valid active set with corresponding hypersphere intersection $\bar{I}$ and $\Theta_{\bar{I}}$ defined as in eqs.~\eqref{UI},~\eqref{thetaI}, it is noted that although $\bar{\boldsymbol{\theta}}\in\bar{I}$, this does not necessarily hold for $\hat{\boldsymbol{\theta}}$, as Fig.~\ref{figure1b} indicates, since $\hat{\boldsymbol{\theta}}\in\Theta_{\bar{I}}$, with $\Theta_{\bar{I}}\not\subset\bar{I}$.}
\end{equation}

Combining eqs.~(\ref{assu2}), (\ref{assu3}), we have that
$$J(\hat{\mathbf{u}},\hat{\boldsymbol{\theta}}) < J(\bar{\mathbf{u}},\bar{\boldsymbol{\theta}})$$
Q.E.D.
{\vskip 10pt}

\subsubsection{Proof of item (B)}

In the sequel, we give two useful Propositions concerning the continuity  of the $F$ and $G$ mappings. In both Propositions, without loss of generality, we consider a valid active set, having $\mathbf{x}_i, i=1,\ldots,k$ as active points, whose corresponding hypersphere intersection is denoted by $I$ and $U_I$, $\Theta_I$ are defined via eqs.~\eqref{UI},~\eqref{thetaI}.
{\vskip 5pt}
{\em Proposition 1:} The mapping $G$ is continuous on $U_I\times \{0\}^{N-k}$.
{\vskip 5pt}
{\em Proof:} To prove that $G$ is continuous in the $N$ variables $u_i$, note that $G$ is a vector field with the resolution by ($l$) scalar fields, written as
$$G=(G_1,\ldots,G_l): U_I\times \{0\}^{N-k} \rightarrow {\cal R}^l$$
where $G_q: U_I\times \{0\}^{N-k} \rightarrow {\cal R}$ is defined as:
\begin{equation}
\label{G-q}
G_q(\mathbf{u})=\frac{ \sum_{i=1}^N u_i \mathbf{x}_i }{\sum_{i=1}^N u_i } \equiv \theta_q,\ \ q=1,\ldots,l
\end{equation}
Since $\{ u_i \rightarrow u_i \mathbf{x}_i \}$ is a continuous function and the sum of continuous functions is also continuous, $G_q$ is also continuous as the quotient of two continuous functions. Under the assumption that $\sum_{i=1}^N u_i>0$, the denominator in eq. (\ref{G-q}) never vanishes. Thus, $G_q$ is well-defined in all cases and it is also continuous. Therefore, $G$ is continuous in its entire domain.  Q.E.D.

{\vskip 5pt}
{\em Proposition 2:} The mapping $F$ is continuous over $I$.
{\vskip 5pt}
{\em Proof:} It suffices to show that $F$ is continuous on the $l$ variables $\theta_q$. $F$ is a vector field with the resolution by ($N$) scalar fields, i.e.,
$$F=(F_1,\ldots,F_N): I \rightarrow U_I$$
where $F_q$ is given by eq. (\ref{globJ1}).

\begin{figure}[htpb!]
\centering
{\includegraphics[width=0.55\textwidth]{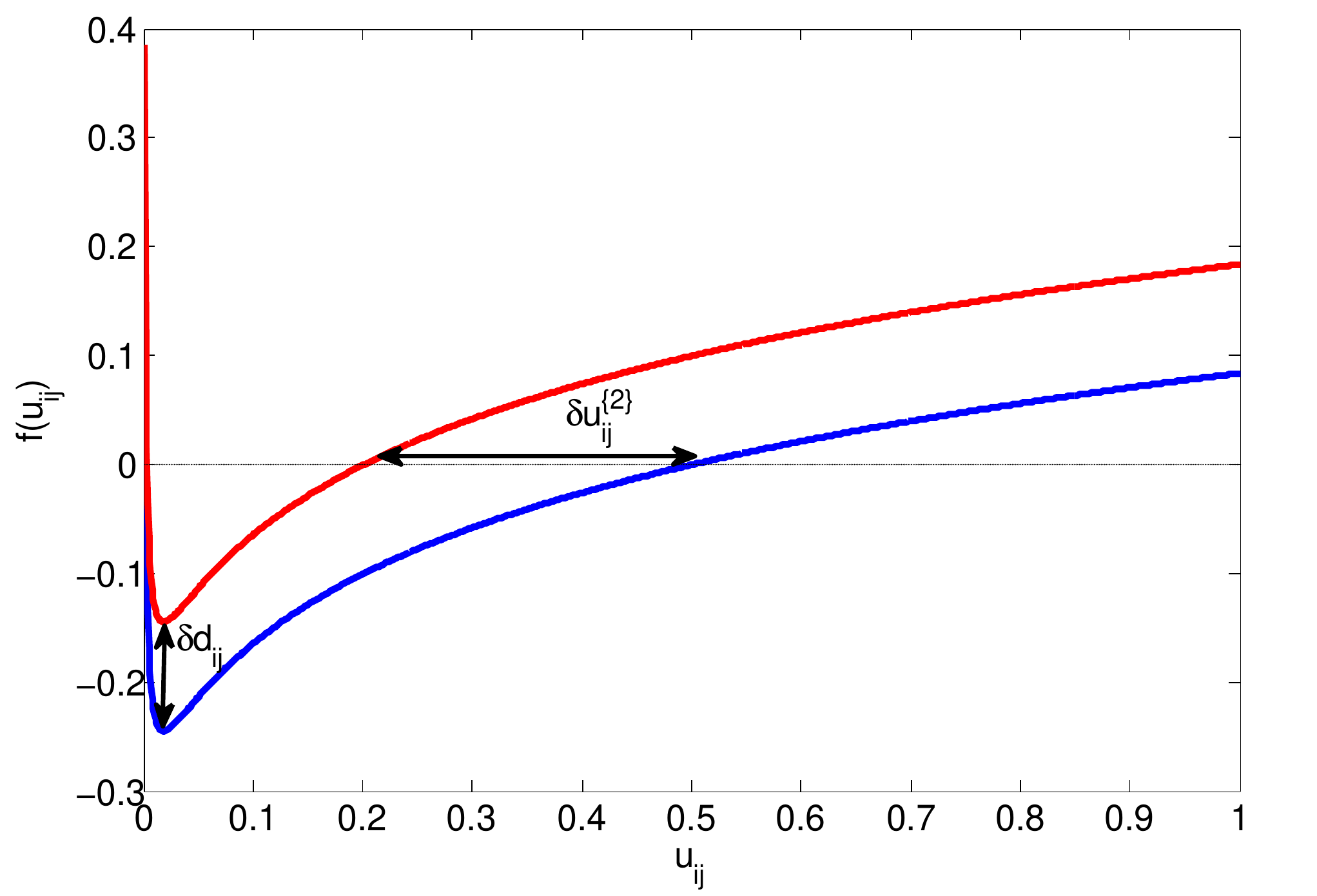}}
\hfil
\centering{\caption{Graphical presentation of the continuity of the mapping $\{d_{ij} \rightarrow u_{ij} \}$. Small variations in $d_{ij}$ cause small variations in $u_{ij}$.}\label{figfu}}
\end{figure}

The mapping $\{ {\mbox{\boldmath $\theta$}} \rightarrow ||{\mathbf{x}}_i-{\mbox{\boldmath $\theta$}}||^2(\equiv d_i)\}$ is continuous. Let us focus on the $u_i$'s, $i=1,\ldots,k$, for which $int({\cal C}_i)$ contributes to the formation of $I$; that is, on $u_i$'s given by the first branch of (\ref{globJ1}). The mapping $\{d_i \rightarrow u_i \}$ is continuous. To see this, note that (since $\gamma$ is constant), the graph of $f(u_i)$ (which is continuous), viewed as a function of $d_i$, is simply shifted upwards or downwards as $d_i$ varies (see fig. \ref{figfu}). Focusing on the rightmost point, $u_i^{\{2\}}$, where the graph intersects the horizontal axis, it is clear that small variations of $d_i$ cause small variations to $u_i^{\{2\}}$, which implies the continuity of $\{d_i \rightarrow u_i \}$ in this case. 

Let us focus next on the $u_i$'s, $i=k+1,\ldots,N$, for which $int({\cal C}_i)$ do not contribute to the formation of $I$; in this case $u_i$ is given by the second branch of (\ref{globJ1}) and the claim follows trivially. Q.E.D.
{\vskip 7pt}

As a direct consequence of Propositions 1 and 2, we have the following lemma.
{\vskip 10pt}
{\em Lemma 5:} $T$ is continuous on $U_I \times I$.
{\vskip 5pt}
{\em Proof:} Recall that $T=T_2 \circ T_1$ and $T_2$ and $T_1$ are defined in terms of $G$ and $F$, respectively (eqs.~\eqref{T1},~\eqref{T2}). $G$ is continuous on $U_I$, as a consequence of Proposition 1, while $F$ is continuous on $I$ from Proposition 2. Thus, $T$ is continuous on $U_I \times I$ as composition of two continuous functions.  Q.E.D.

\subsubsection{Proof of item (C)}

We proceed now to prove that the sequence $(\mathbf{u}^{(t)},\boldsymbol{\theta}^{(t)})\arrowvert_{t=0}^{\infty}$ produced by the SPCM falls in a bounded set.
{\vskip 10pt}
{\em Lemma 6:} Let $(F(\boldsymbol{\theta}^{(0)}),\boldsymbol{\theta}^{(0)})$ be the starting point of the iteration with the SPCM operator $T$, with $\boldsymbol{\theta}^{(0)} \in CH(X)$ and $\mathbf{u}^{(0)}=F(\boldsymbol{\theta}^{(0)})$. Then
$$(\mathbf{u}^{(t)},\boldsymbol{\theta}^{(t)}) \equiv T^t(\mathbf{u}^{(0)},\boldsymbol{\theta}^{(0)}) \in [0,\ 1]^N\times CH(X)$$
{\vskip 5pt}
{\em Proof:} For a given $\boldsymbol{\theta}^{(0)} \in CH(X)$, $\mathbf{u}^{(0)} = F(\boldsymbol{\theta}^{(0)}) \in [0,\ 1]^N$, since $u_i^{(0)} \in [0,\ 1]$ (see eq. (\ref{globJ}) and the argumentation in \cite{Xena16}). Also, $\boldsymbol{\theta}^{(1)}=G(\mathbf{u}^{(0)})$ is computed by eq. (\ref{eq2}), which can be recasted as
$${\mbox{\boldmath $\theta$}}^{(1)} = \sum_{i=1}^N \frac{u_i^{(0)}}{\sum_{i=1}^N u_i^{(0)} } {\mathbf{x}}_i$$
Since $u_i^{(0)} \in [0, 1]$, it easily follows that $0 \leq \frac{u_i^{(0)}}{\sum_{i=1}^N u_i^{(0)} } \leq 1$ and $\sum_{i=1}^N \frac{u_i^{(0)}}{\sum_{i=1}^N u_i^{(0)} } = 1$. Thus ${\mbox{\boldmath $\theta$}}^{(1)} \in CH(X)$. Continuing recursively we have $\mathbf{u}^{(1)} = F(\boldsymbol{\theta}^{(1)}) \in [0,\ 1]^N$ by eq. (\ref{globJ}) and $\boldsymbol{\theta}^{(2)} = G(\mathbf{u}^{(1)}) \in CH(X)$, using the same argumentation as above.
Thus, inductively, we conclude that
$$(\mathbf{u}^{(t)},\boldsymbol{\theta}^{(t)}) \equiv T^t(\mathbf{u}^{(0)},\boldsymbol{\theta}^{(0)}) \in [0,\ 1]^N \times CH(X)$$ Q.E.D.

{\vskip 7pt}
{\em Remark 4:} Note that it is possible to have ${\mbox{\boldmath $\theta$}}^{(0)}$ outside $CH(X)$, yet in a position where at least one $u_i$ is positive. However, computing $\mathbf{u}^{(0)}=F(\boldsymbol{\theta}^{(0)})$ by eq. (\ref{globJ}), the latter will lie in ${\cal M}$ and, as a consequence, $\boldsymbol{\theta}^{(1)}=G(\mathbf{u}^{(0)})$ will lie in $CH(X)$ as it follows by the argumentation given in the proof of Lemma 5.

\subsubsection{Proof of item (D)}


In the sequel, we will prove that the elements of the set $S$ (eq. \ref{S}), for a given valid active set with hyperspheres intersection $I$ (if they exist) are strict local minima of the cost function $J$ and thus the cardinality of $S$ is finite.


The elements of $S$ are the solutions $\mathbf{z}^*=(\mathbf{u}^*,\boldsymbol{\theta}^*) \equiv (u_1^*,\ldots,u_k^*,\theta_1^*,\ldots,\theta_l^*)$ \footnote{Without loss of generality, we assume that the $\mathbf{x}_i$'s, $i=1,\ldots,k$ are the active points of the valid active set under study.} of $\nabla J|_{(\mathbf{u},\boldsymbol{\theta}) }=\boldsymbol{0}$ with $u_i^*$ being the largest of the two solutions of $f_{\boldsymbol{\theta}}(u_i)=0$, $i=1,\ldots,k$. They should satisfy the following equations
\begin{equation}
\label{opt-theta}
2 \sum_{i=1}^k u_i^* (\theta_q^* - x_{iq})=0,\ \ q=1,\ldots,l
\end{equation}
and
\begin{equation}
\label{opt-u}
||\mathbf{x}_i-\boldsymbol{\theta}^*||^2 +\gamma \ln u_i^* + \lambda p u_i^{*^{p-1}}=0,\ \ i=1,\ldots,k
\end{equation}

Then, we have the following lemma.
{\vskip 10pt}
{\em Lemma 7:} The points $\mathbf{z}^*$ that satisfy eqs. (\ref{opt-theta}) and (\ref{opt-u}) (if they exist) are strict local minima of $J$ in the domain $U_I \times I$. Moreover, their number is finite.
{\vskip 5pt}
{\em Proof:}
{\vskip 5pt}
In order to prove that $\mathbf{z}^*$ are local minima we need to prove that the Hessian matrix of $J$ computed at $\mathbf{z}^*$, $H_{\mathbf{z}^*}$, is positive definite over a small region around $\mathbf{z}^*$. It is 
\begin{equation}
\label{H*}
H_{\mathbf{z}^*}= \left [
\begin{array}{cccccccc}
g_1^* & 0 &    & 0 & 2(\theta_1^*-x_{11}) & 2(\theta_2^*-x_{12}) &      & 2(\theta_l^*-x_{1l}) \\
0 & g_2^* &    & 0 & 2(\theta_1^*-x_{21}) & 2(\theta_2^*-x_{22}) &      & 2(\theta_l^*-x_{2l}) \\
\vdots & \vdots & \ddots   & \vdots & \vdots & \vdots          & \ddots    & \vdots \\
0 & 0 &     & g_k^* & 2(\theta_1^*-x_{k1}) & 2(\theta_2^*-x_{k2}) &      & 2(\theta_l^*-x_{kl}) \\
2(\theta_1^*-x_{11}) & 2(\theta_1^*-x_{21}) &      & 2(\theta_1^*-x_{k1}) & 2 \sum_{i=1}^k u_i^* & 0 &     & 0 \\
2(\theta_2^*-x_{12}) & 2(\theta_2^*-x_{22}) &      & 2(\theta_2^*-x_{k2}) & 0 & 2 \sum_{i=1}^k u_i^* &     & 0 \\
\vdots & \vdots & \ddots   & \vdots & \vdots & \vdots          &  \ddots   & \vdots \\
2(\theta_l^*-x_{1l}) & 2(\theta_l^*-x_{2l}) &  \ldots    & 2(\theta_l^*-x_{kl}) & 0 & 0 & \ldots & 2 \sum_{i=1}^k u_i^*
\end{array} \right ]
\end{equation}
where 
\begin{equation}
\label{gi}
g_i^*=\gamma u_i^{*^{-1}} -\lambda p(1-p) u_i^{*^{p-2}},\ i=1,\ldots,k
\end{equation}
 Let $\mathbf{z}'=(\mathbf{u}',\boldsymbol{\theta}') \equiv (u_1',\ldots,u_k',\theta_1',\ldots,\theta_\ell')$ be a point in $U_I \times I$ that is close to $\mathbf{z}^*$. More specifically, let $u_1',\ldots, u_k'$ be close to $u_1^*,\ldots,u_k^*$, respectively, so that 
\begin{equation}
\label{diste}
||\boldsymbol{\theta}^* - \frac{\sum_{i=1}^k u_i' \mathbf{x}_i}{\sum_{i=1}^k u_i'}||<\varepsilon
\end{equation}

After some straightforward algebraic operations it follows that
\begin{equation}
\label{pos-def}
\mathbf{z}'^T H_{\mathbf{z}^*} \mathbf{z}' = 2||\boldsymbol{\theta}'||^2 \sum_{i=1}^k u_i^* + 4 \sum_{i=1}^k u_i' \boldsymbol{\theta}'^T(\boldsymbol{\theta}^*-\mathbf{x}_i) + \sum_{i=1}^k u_i'^2 g_i^*
\end{equation}
It is easy to verify that $\sum_{i=1}^k u_i' \boldsymbol{\theta}'^T(\boldsymbol{\theta}^*-\mathbf{x}_i) = \sum_{i=1}^k u_i' \boldsymbol{\theta}'^T (\boldsymbol{\theta}^* - \frac{\sum_{i=1}^k u_i' \mathbf{x}_i}{\sum_{i=1}^k u_i'} ) \geq - \sum_{i=1}^k u_i' ||\boldsymbol{\theta}'|| \varepsilon$.

Utilizing the fact that $u_i>u^{min} \equiv (\frac{\lambda (1-p)}{\gamma})^{1/(1-p)}$, $i=1,\ldots,k$, for the second appearance of $u_i^*$ in the right hand side of (\ref{gi}), it turns out that
$g_i^* \geq \frac{(1-p)\gamma}{u_i^*}$.

Combining the last two inequalities with eq. (\ref{pos-def}), it follows that
\begin{equation}
\label{pos-def1}
\mathbf{z}'^T H_{\mathbf{z}^*} \mathbf{z}' \geq 2 \sum_{i=1}^k u_i^* ||\boldsymbol{\theta}'||^2 - 4 \sum_{i=1}^k u_i' ||\boldsymbol{\theta}'|| \varepsilon + (1-p)\gamma \sum_{i=1}^k \frac{u_i'^2}{u_i^*} \equiv \phi(||\boldsymbol{\theta}'||)
\end{equation}
Since $\sum_{i=1}^k u_i^*>0$, the second degree polynomial $\phi(||\boldsymbol{\theta}'||)$ becomes positive if and only if its discriminant 
\begin{equation}
\label{discri}
\Delta=8 [ 2 \varepsilon^2 (\sum_{i=1}^k u_i')^2 - (1-p)\gamma \sum_{i=1}^k u_i^* \sum_{i=1}^k \frac{u_i'^2}{u_i^*}]
\end{equation}
is negative. But, from Proposition A2 in Appendix, it is
$$(\sum_{i=1}^k u_i')^2 \leq \sum_{i=1}^k u_i^* \sum_{i=1}^k \frac{u_i'^2}{u_i^*}$$
Also, choosing $\varepsilon< \frac{1}{2} \sqrt{ \frac{(1-p)\gamma}{2} }$, we have that $\Delta$ is negative. As a consequence and due to the continuity of $J$ in $U_I \times I$, $\varepsilon$ defines a region around $\mathbf{z}^*$, for which $\mathbf{z}'^T H_{\mathbf{z}^*} \mathbf{z}' > 0$. Thus $\mathbf{z}^*$ is a strict local minimum.

In addition, since the domain $U_I \times I$ is bounded, it easily follows that the number of strict local minima is finite. Q.E.D.

{\em Remark 5:} It can be shown that in the specific case where (a) $\frac{\gamma}{\bar{\gamma}} < \frac{1}{p} e^{(1-p)^2/2}$ and (b) $K$ in eq. (\ref{lambda}) is chosen in the range $[\frac{\gamma}{\bar{\gamma}} p e^{2-\frac{(1+p)^2}{2}},\ pe^{2(1-p)}]$, then the set $S_q$ (eq. \eqref{S}) that corresponds to each valid active set $X_q$ has one element at the most. The proof of this fact follows the line of proof of lemma 7, with the difference that $\varepsilon$ in eqs. (\ref{diste}), (\ref{pos-def1}) and (\ref{discri}) is replaced by $R$ (since the maximum possible distance between two points in the (nonempty) intersection of hyperspheres of distance $R$, is equal to $R$). Then, the conditions (a) and (b) above follow from the requirement to have $2R^2 < (1-p) \gamma$, in order to have negative discriminant $\Delta$. Utilizing eq. (\ref{lambda}) in the previous requirement it follows that $K > \frac{\gamma}{\bar{\gamma}} p e^{2-\frac{(1+p)^2}{2}}$. Taking into account that $K < p e^{2(1-p)}$ (Proposition A1), condition (a) results from the requirement to have $\frac{\gamma}{\bar{\gamma}} p e^{2-\frac{(1+p)^2}{2}} < p e^{2(1-p)}$.
{\vskip 10pt}
In the sequel we denote by $Y_{\mathbf{z}^*}$ a region around a point $\mathbf{z}^*$ in the set $S_q$ corresponding to a valid active set $X_q$, where $J$ is convex. $Y_{\mathbf{z}^*}$ will be called as a {\em valley} around $\mathbf{z}^*$ (such a region always exists, as shown in proposition A3).

Having completed the proof of the prerequisites (A)-(D) and before we proceed any further, some remarks are in order.

{\em Remark 6:} Although $J$ is well defined in $[0,1]^N \times {\cal R}^l$, there are several regions in the landscape of $J(\mathbf{u},\boldsymbol{\theta})$ that are not accessible by the algorithm. For example, some positions $(\mathbf{u},\boldsymbol{\theta})$ where $u_i < u^{min}$ and those where $\boldsymbol{\theta}$ is expressed through eq. (\ref{theta}) with coefficients $u_i$ less that $u^{min}$, are not accessible by the algorithm.

{\em Remark 7:} It is highlighted again the fact that a certain set of active points $X_q$, with corresponding (nonempty) union of hyperspheres $I_q$ and $U_{I_q}$, $\Theta_{I_q}$ as defined by eqs. (\ref{UI}) and (\ref{thetaI}), respectively, may have no local minima of $J$ in $U_{I_q} \times I_q$ that are accessible by $T$. Equivalently, this means that the solution set $S_q$ (see Lemma 3) corresponding to $X_q$ is empty.

\begin{figure}[htpb!]
\centering
\subfloat[]{\includegraphics[width=0.48\textwidth]{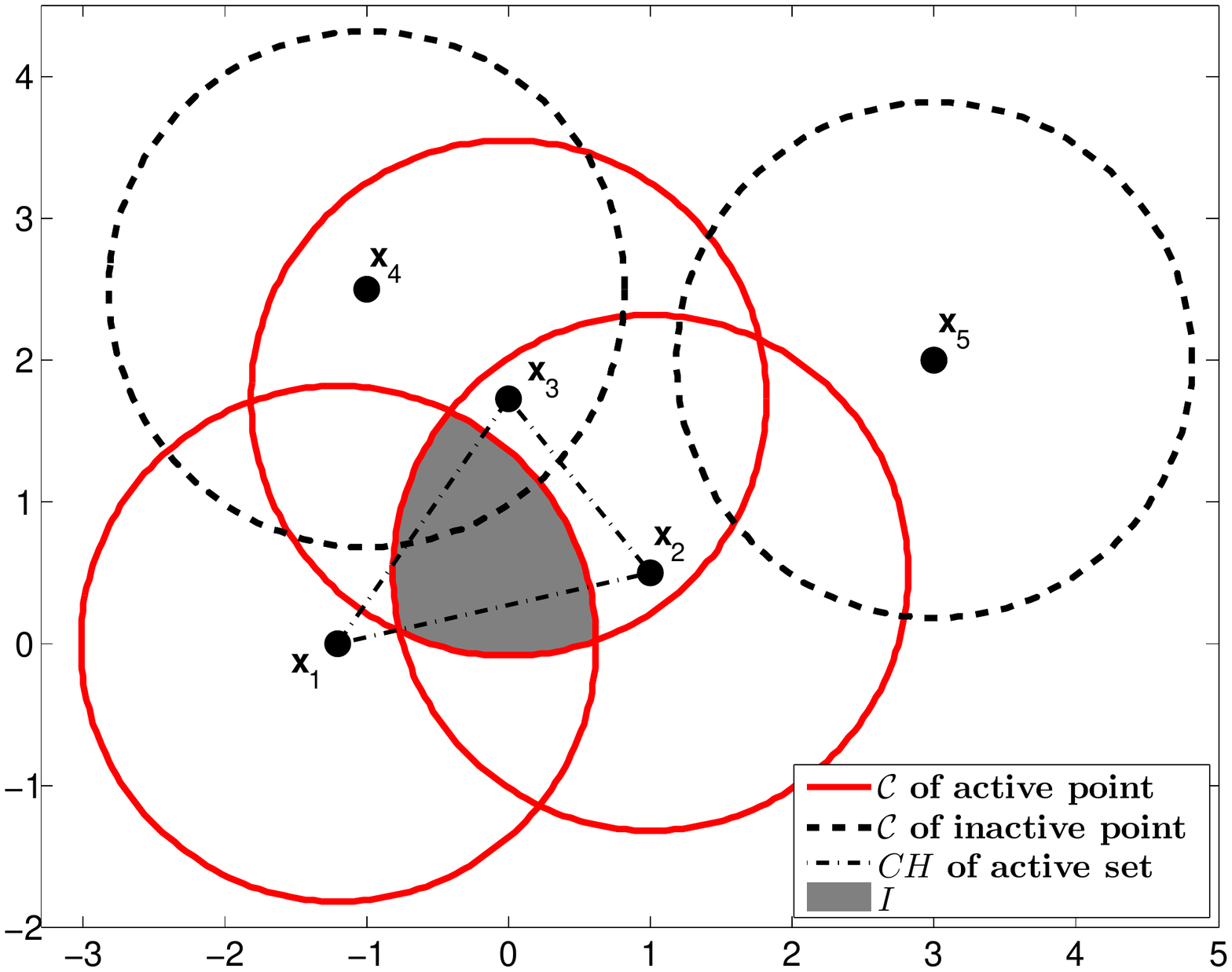}\label{figure2a}}
\hfil
\centering
\subfloat[]{\includegraphics[width=0.48\textwidth]{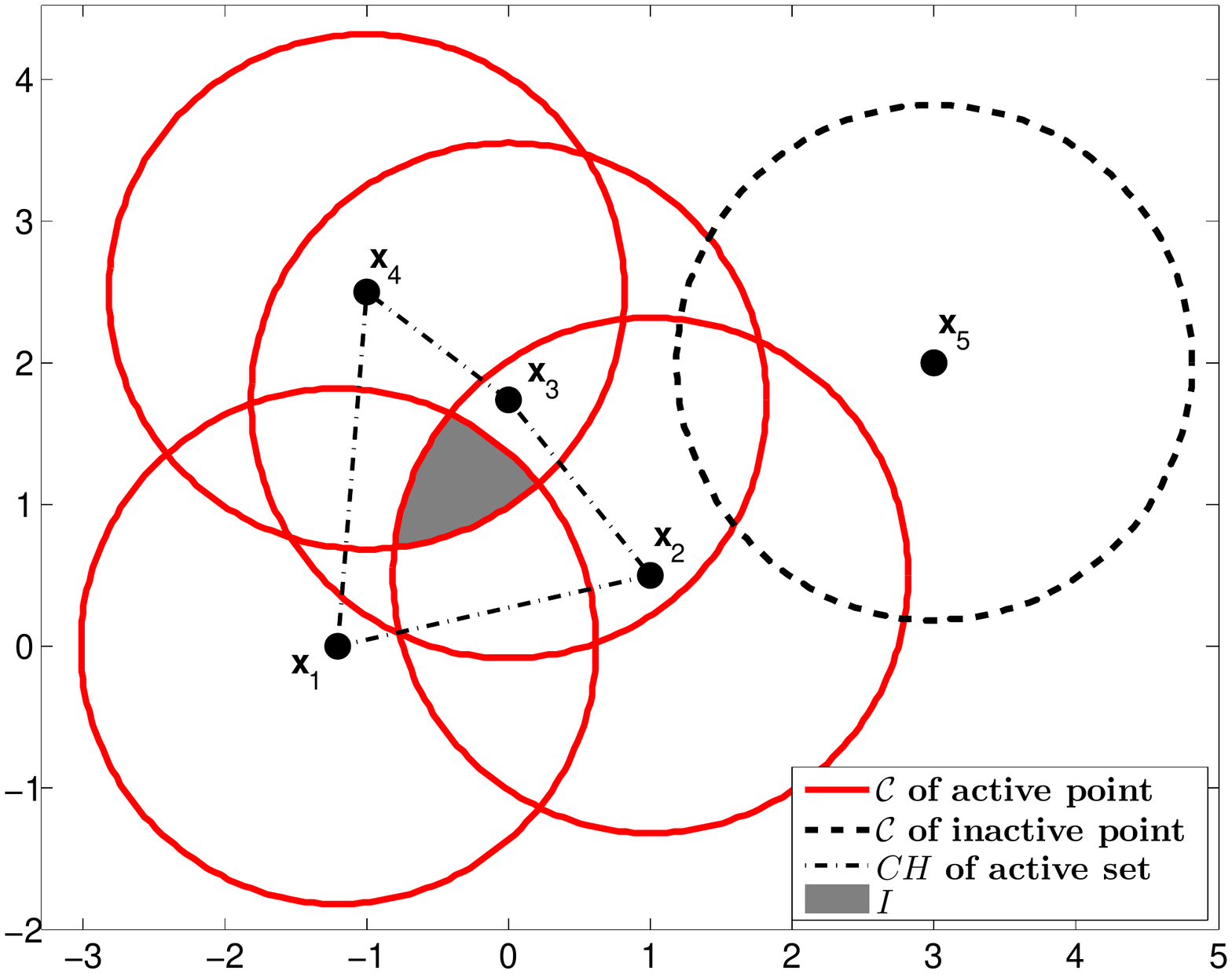}\label{figure2b}}
\hfil
\centering{\caption{(a) An active set of $k=3$ points where $(I \cap (\cap_{i:\ u_i=0} ext({\cal C}_i)))\not\equiv I$ and (b) an active set of $k=4$ points where $(I \cap (\cap_{i:\ u_i=0} ext({\cal C}_i)))\equiv I$}\label{defi2}}
\end{figure}

We prove next the following lemma.
{\vskip 10pt}
{\em Lemma 8:} There exists at least one valid active set $X_q$ (with $I_q \not= \emptyset$) for which there exists at least one local minimum $(\mathbf{u}_{q_r}^*,\boldsymbol{\theta}_{q_r}^*)$, with $\boldsymbol{\theta}_{q_r}^* \in I_q \cap (\cap_{i:\ u_i=0} ext({\cal C}_i))$ \footnote{Note that $\boldsymbol{\theta}_{q_r}^* \in \Theta_{I_q}$ due to the definition of the latter set from eq. (\ref{thetaI}).}.
{\vskip 5pt}
{\em Proof:} Suppose on the contrary that for all possible active sets $X_q$, there is no local minimum $(\mathbf{u}_{q_r}^*,\boldsymbol{\theta}_{q_r}^*)$ with $\boldsymbol{\theta}_{q_r}^* \in I_q \cap (\cap_{i:\ u_i=0}^N ext({\cal C}_i))$ (see fig.~\ref{defi2}). Equivalently, this means that the solution sets $S_q$ for all valid active sets are empty. Then from lemma 3 we have that if at a certain iteration $t_1$, $\boldsymbol{\theta}(t_1)$ belongs to the intersection $I_q$ of a certain active set $X_q$, the algorithm may move $\boldsymbol{\theta}(t)$ ($t>t_1$) to other positions in $I_q$ that always strictly decrease the value of $J$. Since $J$ is bounded below (due to the fact that $\mathbf{u} \in [0,\ 1]^N$ and $\boldsymbol{\theta} \in CH(X)$) it follows that $\boldsymbol{\theta}$ will leave $I_q$ at a certain iteration. In addition, lemma 4 secures the decrease of the value of $J$ as we move from one hypersphere intersection to another (or, equivalently, from one active set to another). Thus, the algorithm will always move $(\mathbf{u}(t),\boldsymbol{\theta}(t))$ from one position to another in the domain $[0,\ 1]^N \times CH(X)$, without converging to any one of them, while, at the same time the value of $J$ decreases from iteration to iteration.

Assuming that at a specific iteration $t'$, $\boldsymbol{\theta}(t')$ belongs to a certain $I_q$, then, due to the continuity of $J$ in $I_q$, there exists a region $V(t')$ around $(\mathbf{u}(t'),\boldsymbol{\theta}(t'))$, for which $J(\mathbf{u},\boldsymbol{\theta}) > J(\mathbf{u}(t'+1),\boldsymbol{\theta}(t'+1))$, for $(\mathbf{u},\boldsymbol{\theta}) \in V(t')$. 

From the previous argumentation, it follows that, since the domain where $(\mathbf{u}(t),\boldsymbol{\theta}(t))$ moves is bounded, the regions $V(t)$ (defined as above) will cover the regions of the whole domain that are accessible by $T$. Thus there exists an iteration $t''$ at which the algorithm will visit a point in the region $V(t')$, where $t'$ is a position the algorithm visited before ($t'<t''$). Then, due to the strict decrease of $J$ as SPCM evolves we have that $J(\mathbf{u}(t''),\boldsymbol{\theta}(t''))< J(\mathbf{u}(t'+1),\boldsymbol{\theta}(t'+1)) < J(\mathbf{u}(t'),\boldsymbol{\theta}(t'))$. However, since $(\mathbf{u}(t''),\boldsymbol{\theta}(t'')) \in V(t')$, it follows that 
$J(\mathbf{u}(t''),\boldsymbol{\theta}(t''))> J(\mathbf{u}(t'+1),\boldsymbol{\theta}(t'+1))$, which leads to a contradiction.
Therefore, there exists at least one active set $X_q$  for which there exists at least one local minimum $(\mathbf{u}_{q_r}^*,\boldsymbol{\theta}_{q_r}^*)$, with $\boldsymbol{\theta}_{q_r}^* \in I_q \cap (\cap_{i:\ u_i=0}^N ext({\cal C}_i))$. Q.E.D.

Now we are in the position to state the general theorem concerning the convergence of SPCM.
{\vskip 10pt}
{\em Theorem 2:} Suppose that a data set $X=\{\mathbf{x}_i\in{\cal R}^l, i=1,\ldots,N\}$ is given. Let $J(\mathbf{u},\boldsymbol{\theta})$ be defined as in eq.~\eqref{Jspcm1} for $m=1$, where $(\mathbf{u},\boldsymbol{\theta})\in{\cal M}\times CH(X)$. If $T: {\cal M}\times CH(X) \rightarrow {\cal M}\times CH(X)$ is the operator corresponding to SPCM algorithm, then for any $(\mathbf{u}(0),\boldsymbol{\theta}(0))\in {\cal M}\times CH(X)$ the SPCM converges to one of the points of the set $S_q$ that corresponds to a valid active set $X_q$, $\mathbf{z_{q_r}}^*=(\mathbf{u}_{q_r}^*,\boldsymbol{\theta}_{q_r}^*)$, provided that $\boldsymbol{\theta}_{q_r}^*\in I_q \cap (\cap_{i:\ u_i=0} ext({\cal C}_i))$.
{\vskip 5pt}
{\em Proof:} Following a reasoning similar to that of lemma 8 we have that the regions of the whole space that are accessible by $T$ will eventually be covered by regions $V(t')$ defined as in the proof of lemma 8. Then the algorithm

(i) either will visit a valley $Y_{\mathbf{z_{q_r}}^*}$ in $U_I\times I_q$ around a (strict) local minimum $(\mathbf{u}_{q_r}^*,\boldsymbol{\theta}_{q_r}^*)$ of a certain active set $X_q$ and, as a consequence of theorem 1 (due to (a) the local convexity of $J$ in $Y_{\mathbf{z_{q_r}}^*}$, (b) the monotonic decrease of $J$ with $T$, (c) the continuity of $T$ in the corresponding $U_I \times I$ and (d) the uniqueness of the minimum in this valley) it will converge to it,

(ii) or it will never visit the valley of such a local minimum. This means that the algorithm starts from a $(\mathbf{u}(0),\boldsymbol{\theta}(0))$, whose $J(\mathbf{u}(0),\boldsymbol{\theta}(0))$ is less than the values of $J$ at all local minima. However,  this case can be rejected following exactly the same reasoning with that in the proof of lemma 8.

Therefore, the algorithm will converge to a local minimum $\boldsymbol{\theta}_{q_r}^*$ that corresponds to one of the possible active sets $X_q$ (with $I_q \not= \emptyset$) provided that $\boldsymbol{\theta}_{q_r}^*\in I_q \cap (\cap_{i:\ u_i=0} ext({\cal C}_i))$.  Q.E.D.

\subsection{Fulfilling the Assumption 1}
\label{subsecKpara}
Next, we show how the {\it Assumption 1} requiring that at each iteration of SPCM at least one equation $f(u_i)=0$, $i=1,\ldots,N$ for each cluster $C_j$, $j=1,\ldots,m$ has two solutions, can always be kept valid. In other words, we show that each cluster has at least one data point $\mathbf{x}_i$, $i=1,\ldots,N$ with $u_i>0$ at each iteration. To this end, we will prove that (a) the {\it Assumption 1} is fulfilled at the initial step of SPCM ({\it base case}) and (b) this inductively holds also for each subsequent iteration of the algorithm ({\it induction step}).

(a) {\it Base case}: Taking into account that the initialization of SPCM is defined by the FCM algorithm and in particular eq.~\eqref{initgamma}, it is obvious that initially each cluster $C_j$ with representative $\boldsymbol{\theta}_j$ has at least one data point with $\|\mathbf{x}_i-\boldsymbol{\theta}_j\|^2\leq\gamma_j$. Focusing on a certain cluster $C_j$, let $\mathbf{x}_q$ be the closest to $\boldsymbol{\theta}_j$ data point, where $\boldsymbol{\theta}_j$ denotes the initial (FCM) estimate of the representative of $C_j$. Then, in general, $\|\mathbf{x}_q-\boldsymbol{\theta}_j\|^2<<\gamma_j$. According to Proposition A4 (see Appendix), this data point has $u_{qj}>0$, if $K\leq\frac{\gamma_j}{\bar{\gamma}}pe^{(2-\mu_j)(1-p)}$, where here $\mu_j=\frac{\|\mathbf{x}_q-\boldsymbol{\theta}_j\|^2}{\gamma_j}(<<1)$. In order to fulfill the {\it Assumption 1} for each cluster, $K$ should be chosen such that $K\leq\min\limits_{j=1,\ldots,m} \left[\frac{\gamma_j}{\bar{\gamma}}pe^{(2-\mu_j)(1-p)}\right]$. Also, it is $\min\limits_{j=1,\ldots,m} \left[\frac{\gamma_j}{\bar{\gamma}}pe^{(2-\mu_j)(1-p)}\right]\geq \frac{\bar{\gamma}}{\bar{\gamma}}pe^{(2-\mu_{max})(1-p)}\equiv pe^{(2-\mu_{max})(1-p)}$, where we recall that $\bar{\gamma}=\min\limits_{j=1,\ldots,m} \gamma_j$. Thus, if $K$ is chosen so that $K\leq pe^{(2-\mu_{max})(1-p)}\equiv B(p)$, where $\mu_{max}=\max\limits_{j=1,\ldots,m} \mu_j(<<1)$, the {\it Assumption 1} is satisfied. Note also that $B(p)\leq pe^{2(1-p)}$, thus the condition of Proposition A1 is valid.

\begin{figure}[htpb!]
\centering
{\includegraphics[width=0.7\textwidth]{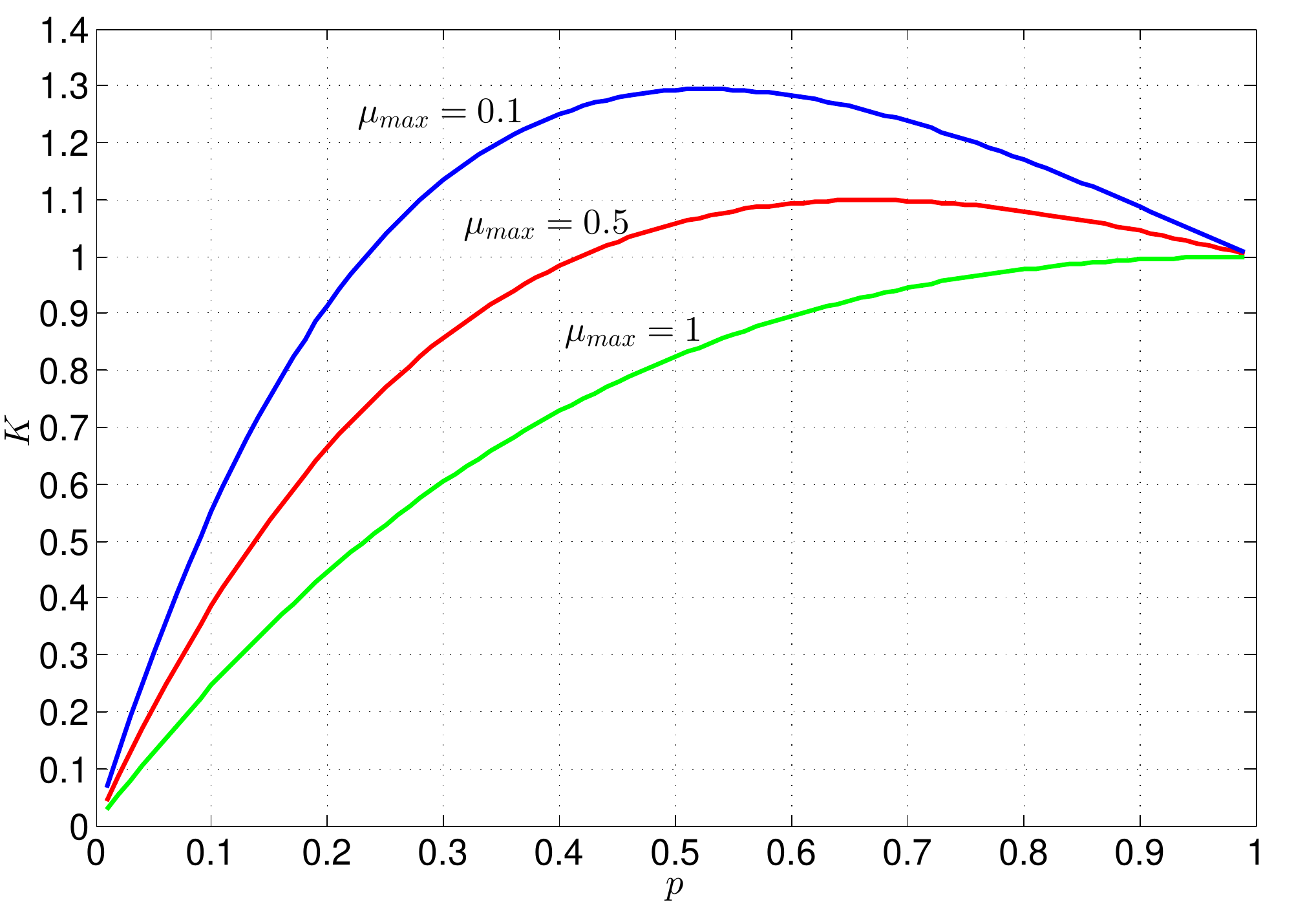}}
\hfil
\centering{\caption{The upper bound $B(p)$ of $K$ with respect to parameter $p$ for different values of $\mu_{max}$, so that each initial cluster has at least one data point with $u>0$.}\label{Kparam}}
\end{figure}

In Fig.~\ref{Kparam}, the upper bound $B(p)$ of $K$ is illustrated with respect to parameter $p$ for different values of $\mu_{max}$, so that each initial cluster has at least one data point with $u>0$. Note that $K=0.9$ is an appropriate value for $p=0.5$ that ensures that the {\it Assumption 1} is fulfilled at the initial step of SPCM (this is the choice made for $K$ in \cite{Xena16}).

(b) {\it Induction step}: Let us focus on a specific cluster $C$ \footnote{For notational convenience, we drop the cluster index $j$ for the rest of this subsection.}. Assume that at iteration $t$, its represenative is $\boldsymbol{\theta}(t)$ and it has a certain set of active points $X^t$ \footnote{We drop the index $q$, in order to lighten the notation. Index $t$ shows the time dependence of the active set corresponding to $C$, as it evolves in time.} with its corresponding nonempty intersection of hyperspheres, denoted by $I^t$. Obviously, it is $CH(X^t)\subseteq(\cup_{i: u_i>0}int({\cal C}_i))$. Taking into consideration that all possible positions of $\boldsymbol{\theta}(t+1)$ lie inside $CH(X^t)$, we have that $\boldsymbol{\theta}(t+1)$ will lie inside $\cup_{i: u_i>0}int({\cal C}_i)$. As a consequence, there exists at least one data point of $X^t$ that will remain active at the next iteration of the algorithm.

As a result, each cluster will have at least one data point $\mathbf{x}_i$, $i=1,\ldots,N$ with $u_i>0$ at each iteration of SPCM.

\section{On the convergence of the PCM$_2$ algorithm}

In \cite{Zhou13} it is proved that the sequence $T^t(U^{(0)},\boldsymbol{\Theta}^{(0)})$ produced by PCM$_2$ terminates to (i) either a local minimum or a saddle point of $J$, or (ii) every convergent subsequence of the above sequence terminates to a local minimum or a saddle point of $J$. This result follows as a direct application of the Zangwill's convergence theorem (\cite{Zang69}). However, viewing PCM$_2$ as a special case of SPCM, we can utilize the convergence results of the latter to establish stronger results for PCM$_2$, compared to those given in \cite{Zhou13}.

Let us be more specific. We focus again to a single $\boldsymbol{\theta}$ and its corresponding $\mathbf{u}=[u_1,\ldots,u_N]^T$ vector. Note that $J_{PCM_2}$ results directly from $J_{SPCM}$, for $\lambda=0$. In this case, the radius $R$ (eq. (\ref{R2})) becomes infinite for any (finite) value of $p$. This means that the convex hull of $X$, $CH(X)$, lies entirely in the intersection of the hyperspheres centered at the data points of $X$. As a consequence, $u_i>0$, for $i=1,\ldots,N$. This implies that the whole $X$ is the active set. Also, note that for $\lambda=0$, $f(u_i)=0$ gives a single positive solution, i.e. $u_i=\exp(-\frac{||\mathbf{x}_i - \boldsymbol{\theta}||^2}{\gamma})$.

Let us define the solution set $S$ for PCM$_2$ as 
$$S_{PCM_2} = \{ (\mathbf{u},\boldsymbol{\theta}) \in [0,\ 1]^N \times CH(X):\ \nabla J|_{(\mathbf{u},\boldsymbol{\theta}) }=\boldsymbol{0} \}$$

The requirements for (i) the decreasing of $J_{PCM_2}$, (ii) the continuity of $T_{PCM_2}$ (the operator that corresponds to PCM$_2$, defined in a fashion similar to $T$) and (iii) the boundness of the sequence produced by PCM$_2$ can be viewed as special cases of Lemmas 3, 5 and 6, respectively, where $U_I \times I$ is replaced by $[0,\ 1]^N \times CH(X)$ \footnote{The only slight difference compared to SPCM concerns the establishment of requirement (i). Specifically, in the proof of Lemma 1 in (eq.~\eqref{contradi-k}), it turns out that for PCM$_2$, it is $\kappa_s=-\infty$, which still contradicts the fact that $\kappa_s$ is finite. Also, in (\ref{contradi-t}) in the same proof it results that $\tau_s \leq 0$, which gives also a contradiction.}.
Then Theorem A1 (see Appendix) guarantees that there exist fixed points for $T_{PCM_2}$ and lemma 7 proves that these are strict local minima of $J_{PCM_2}$ \footnote{The only thing that is differentiated in the PCM$_2$ case is that $g_i^*=\frac{\gamma}{u_i^*}$. As a consequence, $\varepsilon$ is chosen as $\varepsilon<\frac{1}{2}\sqrt{\frac{\gamma}{2}}$.}.
Finally, in correspondance with SPCM, the following theorem can be established for PCM$_2$.
{\vskip 10pt}
{\em Theorem 3:} Suppose that a data set $X=\{ {\mathbf{x}}_i \in {\cal R}^l,\ i=1,\ldots,N \}$ is given. Let 
$J_{PCM_2}(\mathbf{u},\boldsymbol{\theta})$ be defined by eq. (\ref{Jpcm2}) for $m=1$, where $(\mathbf{u},\boldsymbol{\theta}) \in [0,\ 1]^N \times CH(X)$. If $T_{PCM_2}: [0,\ 1]^N \times CH(X) \rightarrow [0,\ 1]^N \times CH(X)$ is the operator corresponding to the PCM$_2$ algorithm, then for any $(\mathbf{u}^{(0)},\boldsymbol{\theta}^{(0)}) \in [0,\ 1]^N \times CH(X)$, the PCM$_2$ algorithm converges to a fixed point of $T$ (which is a local minimum of $J_{PCM_2}$).

\section{Conclusion}
In this paper, a convergence proof for the recently proposed sparse possibilistic c-means (SPCM) algorithm is conducted. The main source of difficulty in the provided SPCM convergence analysis, compared to those given for previous possibilistic algorithms, relies on the updating of the degrees of compatibility, which are not given in closed form and are computed via a two-branch expression. In the present paper, it is shown that the iterative sequence generated by SPCM coverges to a local minimum (fixed point) of its accosiated cost function $J_{SPCM}$. Finally, the above analysis for SPCM has been applied to the case of PCM$_2$ (\cite{Kris96}) and gave much stronger convergence results compared to those provided in \cite{Zhou13}. 

\appendix
{\em Proposition A1:} If $K < pe^{2(1-p)}$, then $R_j>0$.
{\vskip 5pt}
{\em Proof:} Substituting $\lambda$  from eq. (\ref{lambda}) into the definition of $R_j^2$ from eq. (\ref{R2}) and after some manipulations, we have
$$R_j^2 = \frac{\gamma_j}{1-p} \left( -\ln \frac{\bar{\gamma}}{\gamma_j} - \ln \frac{K}{e^{2-p}} - p \right)$$
or, since $\frac{\bar{\gamma}}{\gamma_j}<1$
$$R_j^2 \geq \frac{\gamma_j}{1-p} \left( - \ln \frac{K}{e^{2-p}} - p \right)$$
Straightforward operations show that the positivity of the quantity in parenthesis is equivalent to the hypothesis condition $K<p e^{2(1-p)}$. Q.E.D.
{\vskip 10pt}
{\em Proposition A2:} It is $(\sum_{i=1}^k u_i')^2 \leq \sum_{i=1}^k u_i \sum_{i=1}^k \frac{u_i'^2}{u_i}$, for $u_i, u_i'>0$, $i=1,\ldots,k$.
{\vskip 5pt}
{\em Proof:} It is
$$(\sum_{i=1}^k u_i')^2 \leq \sum_{i=1}^k u_i \sum_{i=1}^k \frac{u_i'^2}{u_i} \Leftrightarrow  \sum_{i=1}^k u_i'^2 +2 \sum_{i=1}^k \sum_{j=i+1}^k u_i' u_j' \leq \sum_{i=1}^k u_i'^2 + \sum_{i=1}^k \sum_{j=1}^k \frac{u_i}{u_j} u_j'^2 \Leftrightarrow$$

$$\sum_{i=1}^k \sum_{j=i+1}^k (\frac{u_i}{u_j} u_j'^2 + \frac{u_j}{u_i} u_i'^2 -2 u_i' u_j') \geq 0 \Leftrightarrow \sum_{i=1}^k \sum_{j=i+1}^k \frac{(u_i u_j' - u_j u_i')^2}{u_i u_j} \geq 0$$
which obviously holds.   Q.E.D.
{\vskip 10pt}

{\em Proposition A3:} Let $\mathbf{z}^*=(\mathbf{u}^*,\boldsymbol{\theta}^*)\in S_q$ corresponding to a certain active set $X_q$. Let also $Y_{\mathbf{z}^*}=Y_{\mathbf{u}} \times Y_{\boldsymbol{\theta}}$ be a set of $(\mathbf{u},\boldsymbol{\theta})$, such that $Y_{\mathbf{u}}=\{\mathbf{u}\in{\cal M}: ||\boldsymbol{\theta}^* - \frac{\sum_{i=1}^k u_i \mathbf{x}_i}{\sum_{i=1}^k u_i}||<\varepsilon\}$ where $\varepsilon<\frac{1}{2}\sqrt{\frac{(1-p)\gamma}{2}}$ and $Y_{\boldsymbol{\theta}}=\{\boldsymbol{\theta}:\boldsymbol{\theta}=\frac{\sum_{i=1}^k u_i \mathbf{x}_i}{\sum_{i=1}^k u_i}, \mathbf{u}\in Y_{\mathbf{u}}\}$. Then (a) $Y_{\mathbf{z}^*}$ is a convex set and (b) $J$ is a convex function over $Y_{\mathbf{z}^*}$.
{\vskip 5pt}
{\em Proof:} (a) Since the domain $Y_{\mathbf{u}}$ of $\mathbf{u}$ is a cartesian product of closed one-dimensional intervals, it is convex. In addition, the set $Y_{\boldsymbol{\theta}}$ is also convex by its definition. Thus $Y_{\mathbf{z}^*}$ is convex.

(b) We prove that for any $\mathbf{z} \in Y$, it is $\mathbf{z}'^T H_{ \mathbf{z}} \mathbf{z}' >0$, $\forall \mathbf{z}' \in Y$. Following a reasoning similar to that in Lemma 7, we end up with the following inequality (with corresponds to eq. \eqref{pos-def1})
\begin{equation}
\label{pos-def2}
\mathbf{z}'^T H_{\mathbf{z}} \mathbf{z}' \geq 2 \sum_{i=1}^k u_i ||\boldsymbol{\theta}'||^2 - 4 \sum_{i=1}^k u_i' ||\boldsymbol{\theta}'|| (2\varepsilon) + (1-p)\gamma \sum_{i=1}^k \frac{u_i'^2}{u_i} \equiv \phi(||\boldsymbol{\theta}'||)
\end{equation}
Note that the factor $2\varepsilon$ in the right hand side of the above inequality, results from the fact that this is the maximum possible difference between two elements in $Y_{\boldsymbol{\theta}}$. The discriminant of $\phi(||\boldsymbol{\theta}'||)$ is
\begin{equation}
\label{discri1}
\Delta=8 [ 8 \varepsilon^2 (\sum_{i=1}^k u_i')^2 - (1-p)\gamma \sum_{i=1}^k u_i^* \sum_{i=1}^k \frac{u_i'^2}{u_i^*}]
\end{equation}
Proposition A2 and the choice of $\varepsilon$ guarantee that $\Delta$ is negative, which implies that $\mathbf{z}'^T H_{\mathbf{z}} \mathbf{z}'>0$ and as a consequence $J$ is convex over $Y_{\mathbf{z}^*}$. Q.E.D.

{\em Proposition A4:} A data point $\mathbf{x}$ has $u>0$ with respect to a cluster $C$ with representative $\boldsymbol{\theta}$ and parameter $\gamma$ or, equivalently, $f(u)=0$ has solution(s), if $K\leq\frac{\gamma}{\bar{\gamma}}pe^{(2-\mu)(1-p)}$, where $\mu=\frac{\|\mathbf{x}-\boldsymbol{\theta}\|^2}{\gamma}$.
{\vskip 5pt}
{\em Proof:} According to eq.~\eqref{R2}, a data point $\mathbf{x}$ has $u>0$ if and only if $\|\mathbf{x}-\boldsymbol{\theta}\|^2\leq R^2\Leftrightarrow \|\mathbf{x}-\boldsymbol{\theta}\|^2\leq\frac{\gamma}{1-p} \left(-\ln\frac{\lambda(1-p)}{\gamma}-p\right)\Leftrightarrow \mu\leq\frac{1}{1-p} \left(-\ln\frac{\lambda(1-p)}{\gamma}-p\right)$, which, using eq.~\eqref{lambda}, gives $\mu\leq\frac{1}{1-p} \left(-\ln\frac{K\bar{\gamma}}{pe^{2-p}\gamma}-p\right)\Leftrightarrow \mu(1-p)\leq-\ln\frac{K\bar{\gamma}}{p\gamma}+2-2p\Leftrightarrow (2-\mu)(1-p)\geq\ln\frac{K\bar{\gamma}}{p\gamma}\Leftrightarrow e^{(2-\mu)(1-p)}\geq\frac{K}{p}\frac{\bar{\gamma}}{\gamma}\Leftrightarrow K\leq\frac{\gamma}{\bar{\gamma}}pe^{(2-\mu)(1-p)}$. Q.E.D.

{\em Theorem A1 (Leray-Schauder-Tychonoff Fixed point theorem, e.g. \cite{Bert89})}: If $X \subset {\cal R}^p$ is nonempty, convex and compact and if $Z:X \rightarrow X$ is a continuous function, there exists $\mathbf{x}^* \in X$, such that $Z(\mathbf{x}^*) = \mathbf{x}^*$ (fixed point).



%

\ifCLASSOPTIONcaptionsoff
  \newpage
\fi

%

%
%
%




\end{document}